\documentclass[letterpaper]{article} 
\usepackage[submission]{aaai25}  
\usepackage{times}  
\usepackage{helvet}  
\usepackage{courier}  
\usepackage[hyphens]{url}  
\usepackage{graphicx} 
\usepackage{natbib}  
\usepackage{caption} 
\usepackage{booktabs} 
\usepackage{algorithm}
\usepackage{algorithmic}

\usepackage{amsfonts} 
\usepackage[table]{xcolor} 
\usepackage{newfloat}
\usepackage{listings}
\DeclareCaptionStyle{ruled}{labelfont=normalfont,labelsep=colon,strut=off} 
\floatstyle{ruled}
\newfloat{listing}{tb}{lst}{}
\floatname{listing}{Listing}
\usepackage{amsmath}

\usepackage[most]{tcolorbox}
\usepackage{enumitem}
\usepackage{lipsum}

\definecolor{myblue}{RGB}{15,50,95}

\title{A Vision-Based Next GPS Location Prediction Model by Reinforcement Learning \\ from Visual Map Feedback}
\author {
    Ruixing Zhang\textsuperscript{\rm 1},
    Yang Zhang\textsuperscript{\rm 1},
    Tongyu Zhu\textsuperscript{\rm 1,\rm 2},
    Leilei Sun\textsuperscript{\rm 1,\rm 2},
    Weifeng Lv\textsuperscript{\rm 1,\rm 2}
}
\affiliations {
    \textsuperscript{\rm 1} the State Key Laboratory of Complex and Critical Software Environment, Beihang University\\
    \textsuperscript{\rm 2}H3I, Beihang University\\
    \textsuperscript{\rm 3}China Mobile Information Technology Center\\
    yyxzhj@buaa.edu.cn, yannazhang@buaa.edu.cn, leileisun@buaa.edu.cn, tongyuzhu@buaa.edu.cn, lwf@buaa.edu.cn
}

\begin{document}

\maketitle

\begin{abstract}
Next Location Prediction is a critical task in the study of human mobility, with wide-ranging applications in urban life.
In practice, when humans attempt to predict the next GPS location, they often visualize the trajectory on a map and reason based on road connectivity and movement trends. However, the vast majority of existing next GPS location prediction models do not reason \textbf{in the way that humans do}. 
Fortunately, the recent development of Vision-Language Models (VLMs) has demonstrated strong capabilities in visual perception and reasoning. 
This opens up a new possibility: by rendering both the road network and trajectory onto an image and leveraging the reasoning abilities of VLMs, we may enable models to perform trajectory inference in a human-like manner.
To explore this idea, we first propose a method called Vision-Guided Location Search (VGLS), which evaluates whether a general VLM is capable of trajectory-based reasoning without modifying any of its internal parameters.
Based on insights from the VGLS results, we further propose our main approach: VLMLocPredictor, which is composed of two stages:
In the first stage, we design Supervised Fine-Tuning (SFT) tasks that help the VLM understand the road network and acquire basic reasoning ability on such visual inputs.
In the second stage, we introduce Reinforcement Learning from Visual Map Feedback, enabling the model to self-improve its next location prediction ability through interaction with the environment. 
Experiments conducted on datasets from four different cities show that our method achieves state-of-the-art (SOTA) performance and exhibits superior cross-city generalization compared to other LLM-based approaches. 
Case studies further demonstrate that, through our framework, VLMs learn to reason effectively over road networks, substantially enhancing their next GPS location prediction capabilities.
The code can be found at \url{https://github.com/Rising0321/VLMLocPredictor}.
\end{abstract}

\section{Introduction}

With the proliferation of smartphones and various IoT devices, vast amounts of GPS data have been collected in recent years, ranging from individual signaling data to vehicle trajectories. To fully unlock the value of this data, next GPS location prediction has become a critical task in spatial-temporal analysis, with broad applications in traffic resource allocation\cite{res1,res2,res3}, epidemic spread forecasting\cite{epic1,epic2,epic3}, and urban governance\cite{urban1,urban2,urban3}.

Next location prediction can be categorized into two sub-tasks: next GPS location prediction and next POI location prediction. For a long time, both tasks shared similar modeling approaches. In early studies, researchers modeled population movement using Markov chains \cite{markov1}, relying on low-order dependencies to predict the next location. With the proposal of Recurrent Neural Networks (RNNs) and Transformers, models such as DeepMove \cite{deepmove} enabled the learning of richer representations by compressing long-range historical trajectory information. Subsequently, the importance of spatial topology was recognized, leading to the integration of graph neural networks that explicitly model road network structures \cite{unimob, getnext}.
More recently, with the rise of large language models (LLMs), researchers have observed that POI data contain rich semantic information. This has led to the development of various LLM-based methods that leverage POI descriptions, achieving strong results in next POI location prediction tasks \cite{agentmove, nextlocllm, jiang}.

However, for next GPS location prediction, most of these methods depart from how humans reason about trajectory. When predicting the next location, humans typically visualize trajectories over a map and reason jointly based on trajectory trends, road network connectivity, and natural assumptions such as drivers usually drive on the shortest road. Current models, in contrast, lack this form of map-based visual reasoning.

Fortunately, the emergence of VLMs has demonstrated strong capabilities in visual understanding and reasoning. This motivates our central hypothesis: \textbf{Can we enable VLMs to reason over visual maps and trajectories, thereby mimicking human-like trajectory inference for next GPS location prediction?} To test this, we first propose a diagnostic framework, Vision-Guided Location Search (VGLS), which evaluates whether a general VLM possesses innate trajectory reasoning ability, without modifying its internal parameters. And our preliminary experiments show that larger VLMs exhibit some capacity for trajectory reasoning, although largely fall behind classic deep learning methods, while smaller VLMs perform near-random. Therefore, transferring the success of VLMs to next GPS location prediction is a non-trivial endeavor, and presents two key challenges:

\begin{itemize}
    \item  \textbf{How can we endow trajectory reasoning abilities to models that lack them?} As shown in our VGLS results, many models exhibit almost no ability to reason over visual maps. Moreover, by the time of submission, most models lack the ability to output the pixel coordinates on the image, making visual reasoning brittle. This calls for a structured approach to progressively instill fine-grained visual grounding and reasoning capabilities.
    \item \textbf{How can we improve a VLM’s ability to reason over visual trajectories in an autonomous way?} Inspired by the success of Reinforcement Learning with Verifiable Reward (RLVR) in models like GPT-o1 and DeepSeek-R1, which enabled large models to surpass supervised methods through autonomous self-improvement, we aim to apply Reinforcement Learning to enhance VLMs' capabilities in a similar way. However, designing task-aligned reward functions and interactive environments that can encourage the model to learn visual reasoning remains an open problem.
\end{itemize}

To address these challenges, we propose VLMLocPredictor, the first framework to enable VLM-based next GPS location prediction using Reinforcement Learning from Visual Map Feedback.
First, we introduce a two-phase SFT process:
In Phase 1, we fine-tune the model using a point localization task, where the VLM learns to locate trajectory points on the visual map, thus acquiring precise spatial grounding.
In Phase 2, we combine the point localization task with chain-of-thought prediction generated by precisely-designed methods to teach the model basic visual trajectory reasoning, without forgetting coordinate understanding.
Next, we propose Reinforcement Learning from Visual Map Feedback. We design a GRPO-based RL framework tailored for visual reasoning, with two key reward functions: Distance Reward and Road Reward and two format reward functions: Format Reward and Distance.
By fine-tuning a Qwen2-VL-2B model using the proposed pipeline, we obtain VLMLocPredictor. Experiments show that our method achieves SOTA performance and demonstrates significantly better cross-city generalization than existing LLM-based approaches.

Our main contributions are summarized as follows:

\begin{itemize}
    
\item  We propose the first VLM-based framework for next GPS location prediction. Our key insight is that, much like how humans reason about future movements by visualizing trajectories on a map, vision-language models can be leveraged to emulate this trajectory reasoning ability.

\item  We design a RL-based post-training framework to improve VLMs' visual trajectory reasoning ability. Our approach enables models to self-improve through interaction with map-based environments. Additionally, we introduce Vision-Guided Location Search (VGLS) as a tool to assess next-point reasoning ability without fine-tuning.

\item  Our method achieves SOTA results on four cities and significantly outperforms other LLM-based approaches in terms of cross-city transferability. Case studies further show that our model has learned to reason over road networks, similar to human reasoning behavior.
\end{itemize}

\section{Preliminaries}

\subsection{Definations}

Definition 1 (\textbf{Trajectory}): Let a trajectory $ \mathbf{T} $ be defined as an ordered sequence of 13 spatial points, denoted as $ \mathbf{T} = \{p_1, p_2, \dots, p_{13}\} $. Each point $ p_i $ ($ i = 1, 2, \dots, 13 $) is represented by its geographic coordinates $ p_i = (lat_i, lon_i) $, where $ lat_i $ and $ lon_i $ correspond to the latitude and longitude values, respectively. 
To ensure consistency and applicability in next location prediction tasks, the trajectory is processed to be uniform with a constant temporal interval $ \Delta t=45s $. 

Definition 2 (\textbf{Next GPS Location Prediction}): This task is to learn a function $ f $ such that: $f(\mathbf{T}_{1:12}) = \hat{p}_{13}$, where $ \hat{p}_{13} $ is the predicted location that approximates the ground-truth $ p_{13} $.

\subsection{Group Relative Policy Optimization (GRPO)}

Group Relative Policy Optimization (GRPO) is a reinforcement learning algorithm tailored for optimizing large models without relying on learned value functions\cite{grpo}. 
Given an input query $q$, the policy $\pi_\theta$ generates a group of $G$ candidate responses $\{o_1, o_2, ..., o_G\}$, each assigned a verifiable reward $r_i = R(q, o_i) \in [0, 1]$. GRPO computes a normalized, group-relative advantage:
\begin{equation}
A_i = \frac{r_i - \text{mean}(r_1, \dots, r_G)}{\text{std}(r_1, \dots, r_G)}.
\end{equation}

These scores reflect how each response compares to others within the same group. The policy is then updated by maximizing the likelihood of better responses while applying a KL penalty to maintain closeness to a reference model $\pi_{\text{ref}}$:
\begin{equation}
\mathcal{L}_{\text{GRPO}} = - \sum_{i=1}^G \left[ \log \pi_\theta(o_i | q) \cdot A_i \right] + \beta \cdot \mathrm{KL}(\pi_\theta(\cdot | q) \| \pi_{\text{ref}}(\cdot | q)),
\end{equation}
which allows the model to improve via verifiable feedback while maintaining response diversity and aligning with prior knowledge.

\section{Preliminary Task: Can General VLMs make next GPS location prediction?}

\subsection{Visual Guided Location Search (VGLS)}
\begin{figure}[ht]
\begin{center}
\centerline{\includegraphics[width=0.9\columnwidth]{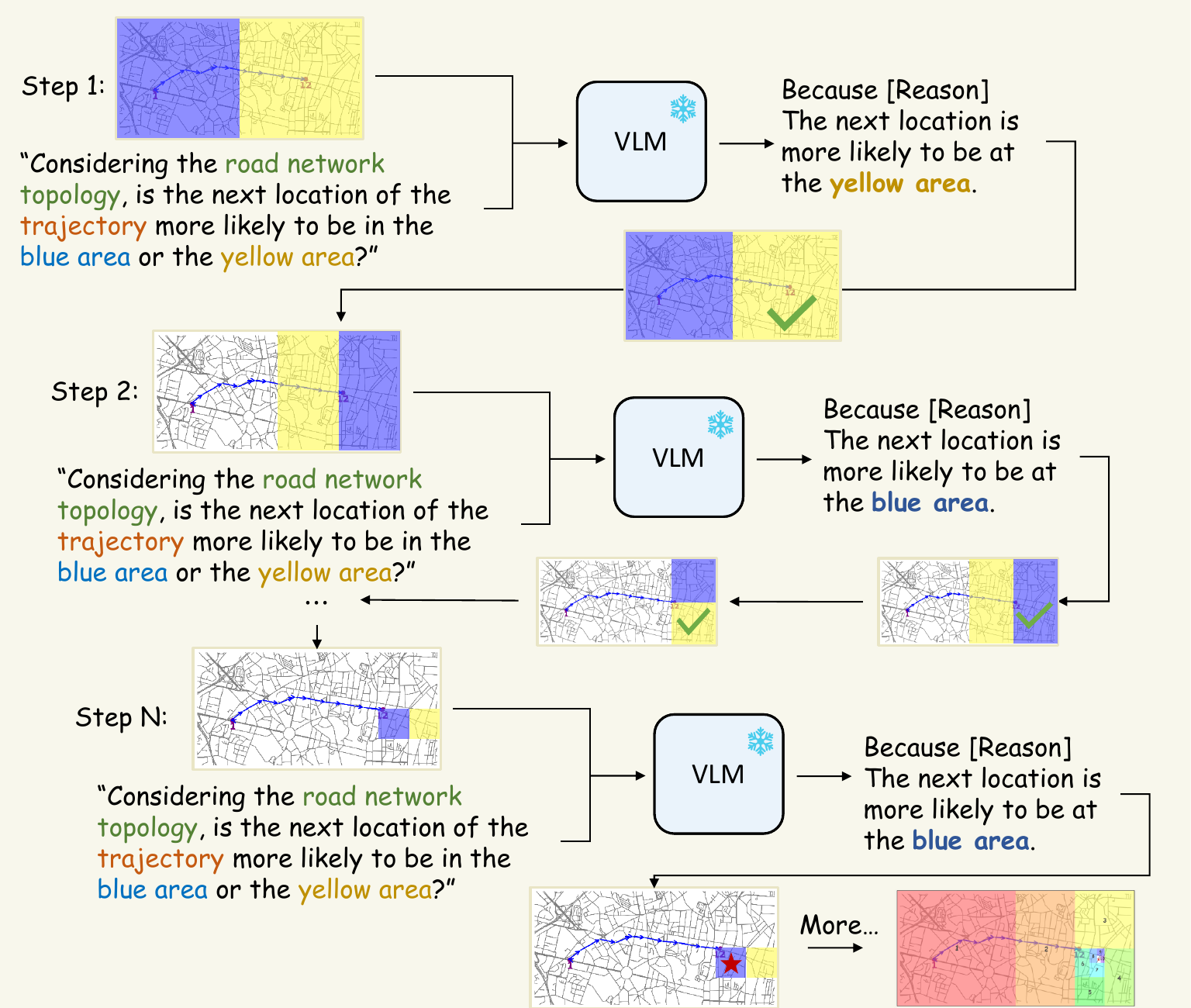}}
\caption{\textbf{An illustration of Visual Guided Location Search.}}
\label{figure1}
\end{center}
\end{figure}
Before delving into the core of this work, we first conduct a preliminary experiment to investigate \textbf{whether general VLMs are capable of next GPS location prediction}. However, evaluating whether a VLM has such capabilities is non-trivial. This difficulty arises because most existing VLMs lack the ability to explicitly pinpoint coordinates on a map. Consequently, designing a method that can assess a VLM’s next location prediction ability presents a significant challenge. Besides, given the goal of this experiment is to assess the visual intelligence of general VLMs, we deliberately avoid fine-tuning or task-specific adaptation methods, as such approaches would contradict the premise of evaluating general capabilities. 

We are inspired by related works that proved VLMs are capable of interpreting specific regions delineated by a circle\cite{circle} and can handle tasks recursively\cite{saining}. Building on these insights, we propose a Visual Guided Location Search (VGLS) mechanism. As shown in Figure \ref{figure1}, the core idea is as follows: we begin by dividing the map into two regions, coloring one half blue and the other half yellow. The VLM is then tasked with determining whether the next trajectory point is more likely located in the blue or yellow region. Take the model identifies the blue region as more likely as an example, the blue region is further subdivided into one half colored blue and the other yellow. This iterative process is repeated for $N$ rounds, progressively narrowing down the feasible region.

Fundamentally, we designed a hierarchical question-answering process. In the first iteration, the model only needs to answer a relatively simple question: which half of the map is more likely to contain the next point. As the process continues, the questions become increasingly challenging. Through this approach, we first designed a method enabling VLMs to perform next GPS location prediction \textbf{without modifying any of their parameters.}

\subsection{Findings}
\begin{figure}[ht]
\begin{center}
\centerline{\includegraphics[width=0.9\columnwidth]{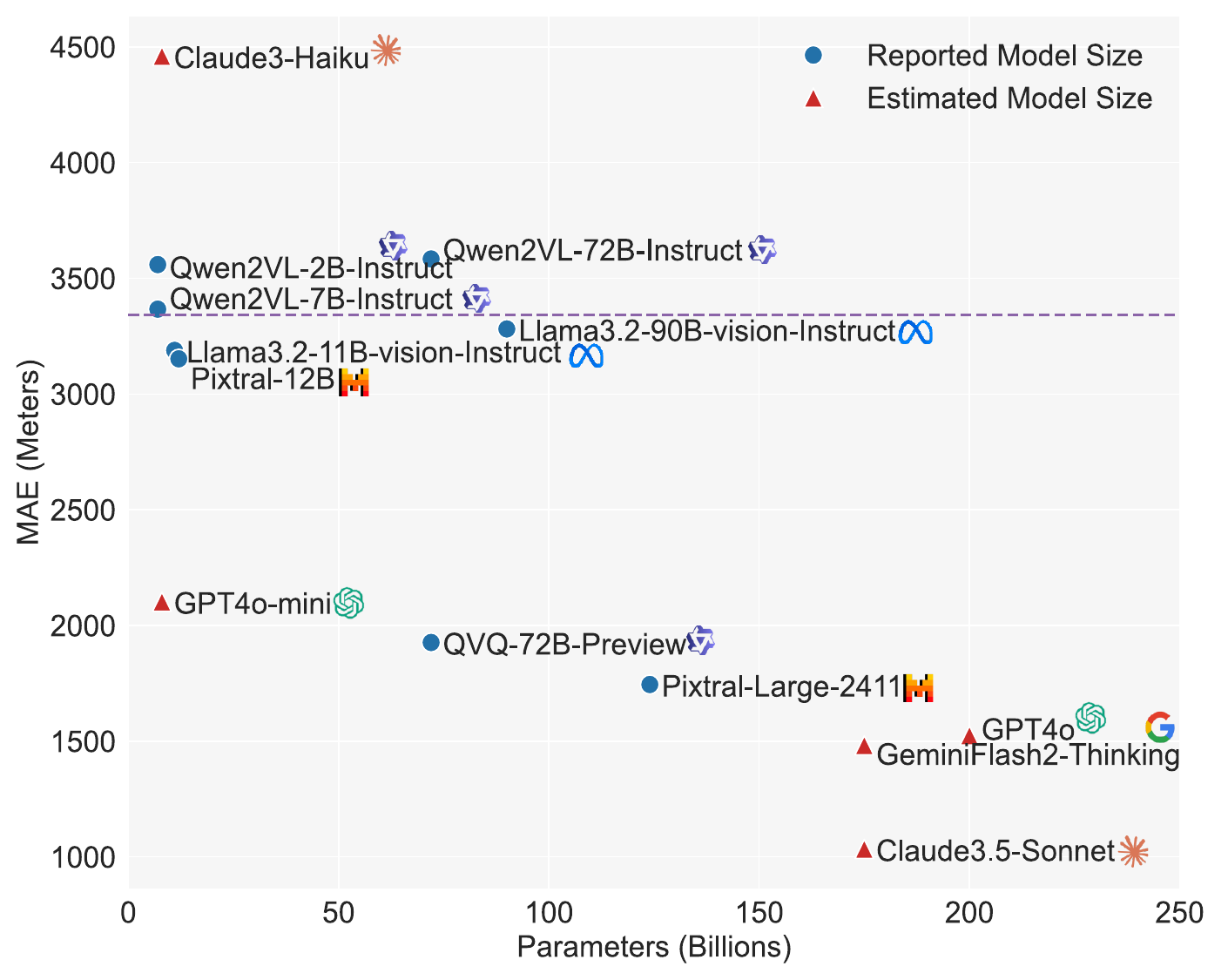}}
\caption{\textbf{Preliminary experiments on chengdu dataset.} }
\label{figure2}
\end{center}
\end{figure}

As shown in Figure \ref{figure2}, the x-axis represents the number of model parameters, while the y-axis indicates the  error between the predicted center point and the ground-truth next location after ten iterations. The dashed line in the figure denotes the error when the model randomly outputs either yellow or blue at each iteration. Due to space limitations, detailed experimental settings please refer to experiment setting and we briefly explain the key findings.:

\begin{itemize}
    \item  Larger models such as GPT-4o, Claude 3.5 exhibit a certain degree of next location prediction capability. This suggests that some general-purpose\textbf{ VLMs can perform next-GPS-point prediction.} To the best of our knowledge, this is the first work to report such a finding.
    \item  Most smaller models, including Qwen2-VL-2B, show performance close to random guessing, indicating a lack of effective next GPS location prediction ability.
    \item  The next GPS location prediction ability of general VLMs has still fallen behind classic baselines(see in the experiment) whose MAE are all below 600 meters.
\end{itemize}
The subsequent parts of our work are primarily built upon the first and second findings above.

\begin{figure*}[ht]
\begin{center}
\centerline{\includegraphics[width=2.2\columnwidth]{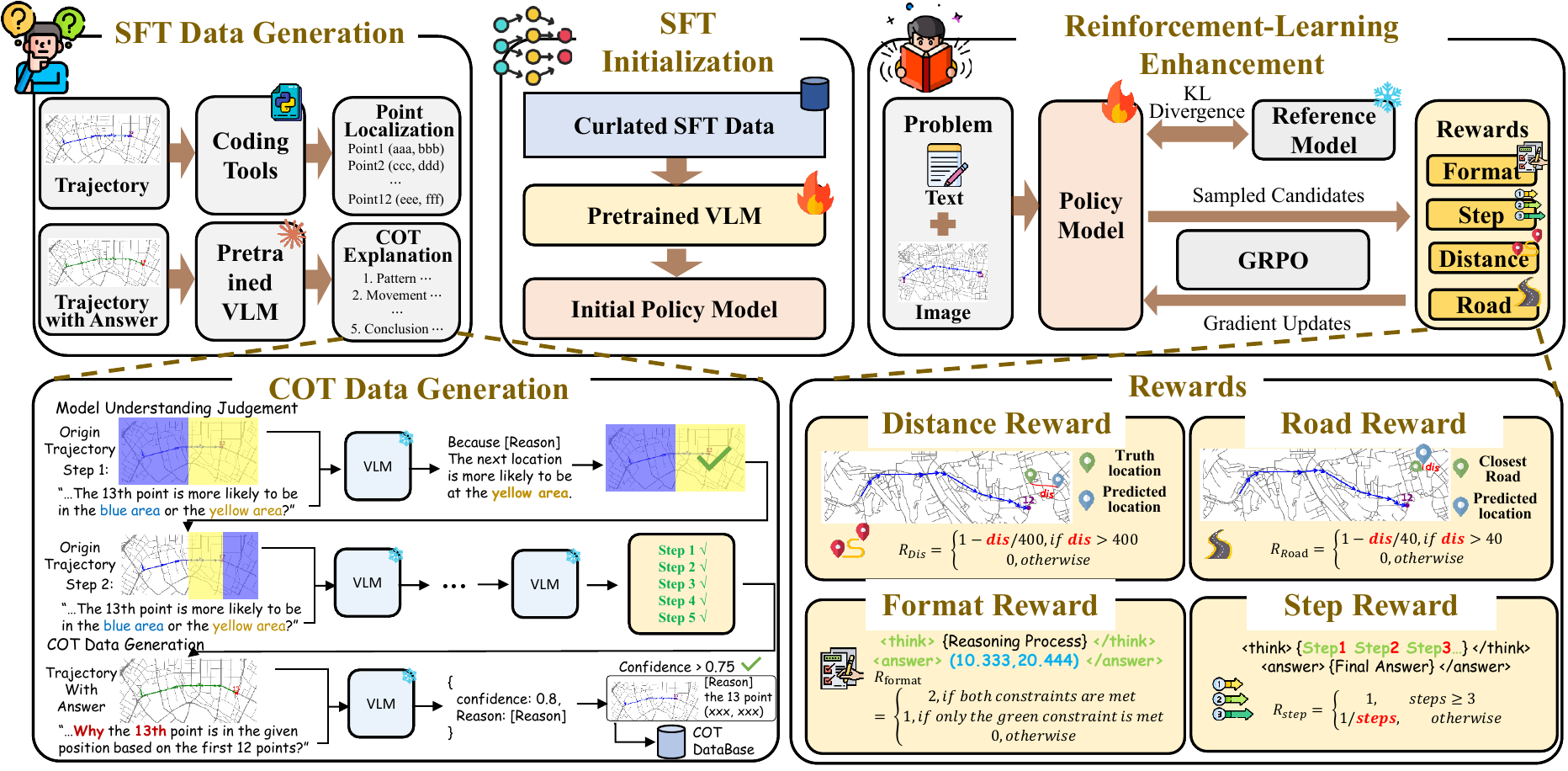}}
\caption{\textbf{The illustration of VLMLocPredictor.} Our approach consists of two stages. In the first stage, we fine-tune a VLM using a carefully constructed dataset, enabling the model to recognize the coordinates of each point on the map and acquire basic trajectory reasoning capabilities. To build the CoT data for SFT, we select samples that pass five rounds of VGLS and ask Claude3.5-Sonet to generate explanations. Only those with confidence scores above 75\% are included. In the second stage, we adopt a reinforcement learning strategy to autonomously enhance the VLM’s ability to predict the next GPS Location.}
\label{figure3}
\end{center}
\end{figure*}

\section{VLMLocPredictor}

In this section, we introduce VLMLocPredictor, a method designed to equip VLMs with next GPS location prediction abilities through Reinforcement Learning from Visual Map Feedback. \textbf{Unlike the binary classification setting in the Preliminary Task, VLMLocPredictor aims to enable the model to directly output the coordinates of the next location point on an image}. The framework is illustrated in Figure \ref{figure3}. To achieve this, we design a two-stage training pipeline:
1. \textbf{SFT-based Initialization}: This stage involves SFT on two tasks, aimed at teaching the VLM to localize points on the image and perform basic trajectory reasoning.
2. \textbf{RL-based Enhancement}: In this stage, we introduce four reward functions and apply GRPO-based RL to enhance the model's reasoning ability in predicting the next GPS point.

\subsection{SFT-based Initialization}
Although RL can be viewed as a means to autonomously enhance an existing capability, if the model lacks the foundational ability, RL may become extremely unexplainable. By the time of this submission, most VLMs cannot output precise coordinates of points on an image and our preliminary experiments showed that some VLMs do not possess visual-based trajectory reasoning ability. Therefore, the goal of the SFT stage is to impart these two fundamental capabilities to the model. To this end, we design two supervised tasks:

\subsubsection{Task 1: Point Localization} Given a trajectory overlaid on an image, the model is required to directly output the coordinate of one specific point. The purpose is to ensure the model accurately understands the visual location of each point on the image. Since accurate next GPS location prediction depends on a clear understanding of the trajectory points, this task serves as the foundation. The dataset used here is denoted as $\mathcal{D}_1$. The construction of this dataset can be finished by common programming tools.

\subsubsection{Task 2: COT Prediction} This task is equivalent to the main problem of predicting the coordinates of the 13th point. Especially, in this task, we additionally require the model to output the chain-of-thought (CoT) that explains how the prediction is made. This task helps the model acquire basic next location reasoning ability, and the corresponding dataset is denoted as $\mathcal{D}_2$. \textbf{However, how to obtain CoT data tailored to the next location prediction task still remains a challenge.}

Based on the findings from the Preliminary Task, Claude 3.5-Sonet demonstrated strong capabilities in visual trajectory reasoning, we aim to use it to generate CoT data. However, to ensure the reliability of the generated CoT, we have to design the following three-step procedure:

\begin{itemize}
    \item We first leverage the VGLS method as a reliability tester. If a trajectory consistently yields correct predictions across the first five iterations of VGLS, we consider the model to have a strong understanding of that trajectory. 
    \item Because Claude 3.5-Sonet cannot directly output the coordinates of a point on the image, we provide it with the full image containing the 13th point and ask it to explain how the next location can be inferred from the first 12 points. Alongside the explanation, the model is also prompted to provide a confidence score for its reasoning.
    \item  We filter and retain only those CoT outputs with a confidence score above 75\%, treating them as high-quality supervision signals. By using the image with 12 trajectory points as the input and concatenating the generated CoT with the true coordinate of the 13th point as the output, we construct a training set for Task 2.
\end{itemize}

Based on these datasets, we design a two-phase SFT pipeline. In the first phase, the model is fine-tuned solely on $\mathcal{D}_1$ to learn point localization ability. In the second phase, we fine-tune on both $\mathcal{D}_1$ and $\mathcal{D}_2$ to instill basic trajectory reasoning while preserving the model’s spatial understanding. This process can be formally expressed as:
\begin{equation}    
\begin{aligned}
\theta_1 =& \arg\max_{\theta \leftarrow \theta_0} \; \mathbb{E}_{(x, y) \sim \mathcal{D}_1} \left[ \log p_\theta(y \mid x) \right] \\
\theta_2 =& \arg\max_{\theta \leftarrow \theta_1} \; \mathbb{E}_{(x, y) \sim \mathcal{D}_1 \cup \mathcal{D}_2} \left[ \log p_\theta(y \mid x) \right] 
\end{aligned}
\end{equation}
which 
$\theta$ denotes the parameters of the VLM, 
$\theta_0$ denotes the pretrained parameters of the VLM, 
$x$ represents the input trajectory with map and prompts, $y$ is the corresponding target coordinates(and CoT data, for task 2), 
and $\theta \leftarrow \theta_0$ denotes the $\theta$ is initialized as $\theta_0$,


\subsection{RL-based Enhancement}

Inspired by RL-based post-training methods, our goal in this stage is to enhance the model’s ability to predict the next GPS location in an autonomous way. Due to the high computational cost of training a value network in  Proximal Policy Optimization-like frameworks, we adopt GRPO, which significantly reduces training overhead during post-training. To guide the learning process, we design four distinct reward functions, each capturing a different aspect of desired model behavior: Distance Reward, Road Reward, Format Reward and Step Reward. We describe each of them in detail below.

\begin{table*}[!ht]
    \centering
    \caption{\textbf{Comparison Results}. We compare our method with six baselines across all datasets. Our approach consistently achieves SOTA results. The bolded scores indicate the best performance, and the underlined scores indicate the second-best.}
    \begin{tabular}{c|rr|rr|rr|rr}
    \toprule
        ~ &  \multicolumn{2}{|c}{\textbf{Chengdu}} & \multicolumn{2}{|c}{\textbf{Porto}} & \multicolumn{2}{|c}{\textbf{Sanfrancisco}}  & \multicolumn{2}{|c}{\textbf{Rome}}  \\ 
        ~ & $\text{MAE}$ & $\text{RMSE}$ & $\text{MAE}$ & $\text{RMSE}$ & $\text{MAE}$ & $\text{RMSE}$ & $\text{MAE}$ & $\text{RMSE}$ \\ 
        \midrule
        \rowcolor{green!20} \multicolumn{9}{c}{Classic Deep Learning Models} \\ \midrule
        DeepMove & 331.464  & 429.775  & 571.015  & 560.972  & 249.929  & 338.508  & \underline{354.001}  & 466.624  \\ 
        GETNext & 330.539  & 424.014  & 563.362  & 584.613  & 268.993  & 365.616  & 388.955  & 497.059  \\ 
        TrajGDM & 439.776  & 512.672  & 681.156  & 671.920  & 311.311  & 397.072  & 410.533  & 493.172  \\ 
        UniMob & 272.154  & \underline{373.764}  & \underline{496.597}  & 522.604  & \underline{216.446}  & \underline{345.373}  & 362.703  & 509.424  \\ 
        \midrule
        \rowcolor{blue!20} \multicolumn{9}{c}{Large Language Model-Based} \\ \midrule
        AgentMove & 589.848  & 842.128  & 1118.883  & 994.936  & 416.731  & 596.207  & 593.628  & 825.680  \\ 
        NextLocLLM & \underline{254.476}  & 429.954  & 571.254  & \underline{484.617}  & 262.483  & 397.306  & 378.920  & \underline{430.894}  \\        \midrule
        \rowcolor{yellow!20} \multicolumn{9}{c}{Vision-Language Model-Based} \\ \midrule
        \textbf{VLMLocPredictor}& \textbf{253.225}  & \textbf{340.708}  & \textbf{452.678}  & \textbf{474.213}  & \textbf{213.233}  & \textbf{333.387}  & \textbf{310.648}  & \textbf{407.737}  \\ 
    \bottomrule
    \label{tab1}
    \end{tabular}
\end{table*}

\textbf{Distance Reward}.
The most critical metric for evaluating the quality of next GPS location prediction is the prediction error. Therefore, we directly incorporate the Euclidean distance as a reward function. Specifically, if the predicted point is more than 400 pixels away from the ground-truth location, the reward is set to 0. Otherwise, the reward is calculated as $R_{\text{dis}} = 1 - \frac{\text{dis}}{400}$.
This design reflects the intuition that predictions with large errors are essentially unusable and should not be rewarded, while more accurate predictions are increasingly incentivized. The Distance reward can be formalized as:
\begin{equation}
R_{\text{dis}} = 
\begin{cases}
1 - \frac{\text{dis}}{400}, & \text{if } \text{dis} \leq 400 \\
0, & \text{otherwise}
\end{cases}
\end{equation}

\textbf{Road Reward}.
Since all the trajectories in our dataset lie on roads, ensuring that the predicted point also falls on the road network is crucial for usability. We find that prompting alone is insufficient to guarantee this behavior in VLMs. Hence, we introduce a Road Reward to encourage on-road predictions.
The reward is calculated based on the Euclidean distance from the predicted point to the nearest road. If this distance exceeds 40 pixels, the reward is 0. Otherwise, it is defined as $R_{\text{Road}} = 1 - \frac{\text{dis}}{40}$. This incentivizes the model to generate predictions that are geographically aligned with the road network.
\begin{equation}
R_{\text{Road}} = 
\begin{cases}
1 - \frac{\text{dis}}{40}, & \text{if } \text{dis} \leq 40 \\
0, & \text{otherwise}
\end{cases}
\end{equation}

\textbf{Format Reward}.
In reinforcement learning-based fine-tuning of large language models, it is common to include a Format Reward to ensure structured outputs. In our setting, we require the model to wrap its reasoning process within $<\text{think}>$ tags and the final answer within $<\text{answer}>$ tags to facilitate downstream parsing. Additionally, the answer must be a coordinate tuple for further processing.
The reward is defined as follows: if the reasoning and answer are properly enclosed in the correct tags, the model receives a reward of 1; if the answer is also correctly formatted as a tuple, the reward is increased to 2. If neither condition is met, the reward is 0. This structured output facilitates both interpretability and automated evaluation.
\begin{equation}
R_{\text{Format}} =
\begin{cases}
2, & \text{if properly enclosed and correctly formatted} \\
1, & \text{if the output is properly enclosed} \\
0, & \text{otherwise}
\end{cases}
\end{equation}

\textbf{Step Reward}.
To further encourage structured reasoning, we introduce a Step Reward that rewards multi-step logical thinking. Specifically, we require the model to generate a step-by-step reasoning trace. If the model outputs three or more clearly distinguishable reasoning steps, the reward is 1. Otherwise, it is linearly decayed based on the number of steps using the formula $R_{\text{Step}} = 1 - \frac{3 - \text{step}}{3}$.
This reward guides the model toward producing more logical, interpretable, and robust reasoning processes.
\begin{equation}
R_{\text{Step}} =
\begin{cases}
1, & \text{if steps} \geq 3 \\
1 - \frac{3 - \text{step}}{3}, & \text{otherwise}
\end{cases}
\end{equation}

Together, these four reward signals serve complementary purposes: Distance and Road Rewards guide geographical correctness, Format Reward ensures output quality and structure, and Step Reward promotes transparent and multi-step reasoning. By integrating them within a GRPO-based framework, we efficiently enhance the model’s ability to reason about and predict the next GPS location on our designed visual maps.

\section{Experiments}

\subsection{Experimental Settings}

\textbf{Datasets.} To evaluate our model’s reasoning ability over visual maps, we construct four visual map-based next location prediction datasets: Chengdu, Porto, San Francisco and Rome. Each dataset contains real-world urban trajectories projected onto static map images. We split the data into training, validation, and test sets with a 7:1:2 ratio. Due to space limitations, detailed descriptions of each dataset can be found in the Appendix.

\textbf{Evaluation Metrics.} We adopt two widely-used metrics for spatial prediction: Mean Average Error(MAE) and Root Mean Square Error(RMSE). Both errors are fundamentally calculated from the Euclidean error between predicted points and the ground-truth. The definition can be found in the appendix. The units are in meters. Because the Format Reward and Step Reward reach 1 after our post-training, we did not show corresponding results.

\textbf{Implementation Details.} We use Qwen2-VL-2B as our base model. All experiments are conducted on four NVIDIA A100 GPUs (80GB each). We do not report results on larger-scale models such as Qwen2.5-VL-3B due to hardware limitations, as RL training is only feasible up to the 2B model size on our infrastructure.

\textbf{Baselines.} We compare our method against six representative baselines, including four deep learning models and two large language model (LLM)-based approaches:
\textbf{DeepMove}\cite{deepmove}: A RNN-transformer hybrid model that captures temporal and spatial patterns in urban mobility.
\textbf{GETNext}\cite{getnext}: A graph-based transformer with multiple losses for next location prediction.
\textbf{TrajGDM}\cite{trajGDM}: A diffusion model that models spatiotemporal dependencies for next location prediction.
\textbf{UniMob}\cite{unimob}: A unified attention-based architecture for trajectory-based prediction tasks.
\textbf{AgentMove}\cite{agentmove}: A recent training-free agent-centric model that simulates goal-driven movement in complex urban environments.
\textbf{NextLocLLM}\cite{nextlocllm}: A method that fine-tunes an LLM to incorporate POI and GPS information for trajectory prediction.

\subsection{Comparision Experimental Results}

To evaluate the effectiveness of our proposed method, we present in Table \ref{tab1} a comparison with six baselines across four datasets. In this experiment, all trainable models are trained on their respective target datasets. The best results are highlighted in bold, while the second-best results are underlined.

We first observe that training-free LLM-based methods, such as AgentMove, perform significantly worse than supervised baselines. Although AgentMove achieves strong results in next POI prediction tasks, it suffers in the next GPS location prediction task. This may be due to the lack of rich semantic information in GPS coordinates compared to POIs, indicating a fundamental difference between the two tasks and the need for different modeling strategies.

Secondly, both LLM-based and classic deep learning methods exhibit reasonable performance after supervised training. However, the performance gap between the two is not substantial, possibly because both have effectively leveraged sequential patterns inherent in trajectory data and may reach the limit of pure sequential data.

In contrast, our method consistently outperforms all baselines across all metrics and all four cities. By incorporating visual maps into the prediction process, our approach further enhances the model’s ability to reason about spatial structures and pushes the limit of next GPS location prediction. These results validate both the efficacy of using visual maps for next GPS location prediction and the design of our proposed learning paradigm.

\subsection{Cross-city Experimental Results}

\begin{table}[h]
    \centering
    \caption{\textbf{Cross-city Experimental Results}. We compare our method with LLM-based models under cross-city scenarios.}
    
    \begin{tabular}{c|rr|rr}
    \toprule
        Target City & \multicolumn{2}{|c}{Chengdu} & \multicolumn{2}{|c}{Porto} \\ 
        \midrule
        Source City & $\text{MAE}$ & $\text{RMSE}$ & $\text{MAE}$ & $\text{RMSE}$ \\  
        \midrule
        \rowcolor{green!20} \multicolumn{5}{c}{AgentMove} \\
        \midrule
        / & 589.848  & 842.128  & 1118.883  & 994.936  \\
        \midrule
        \rowcolor{blue!20} \multicolumn{5}{c}{NextLocLLM} \\
        \midrule
        Chengdu & 254.476  & 429.954  & 885.269  & 1150.359  \\ 
        Porto & 373.449  & 470.615  & 571.254  & 484.617  \\ 
        Rome & 351.592  & 481.091  & 792.534  & 994.676  \\ 
        Sanfrancisco & 337.816  & 464.599  & 848.673  & 1038.538  \\  
        \midrule
        \rowcolor{yellow!20} \multicolumn{5}{c}{\textbf{VLMLocPredictor}} \\
        \midrule
        Chengdu & 253.225  & 340.708  & 651.597  & 832.220  \\ 
        Porto & 368.342  & 431.068  & 452.678  & 474.213  \\ 
        Rome & 314.233  & 397.765  & 590.172  & 795.273  \\ 
        Sanfrancisco & 309.363  & 406.006  & 752.551  & 843.446 \\ 
    \bottomrule
    \end{tabular}
    \label{tab2}
\end{table}

Given that LLM-based methods rely on natural language, they are often expected to generalize across cities. We evaluate cross-city generalization by comparing our method with AgentMove and NextLocLLM in transfer settings.
Specifically, we train models on four cities then test their performance when transferred to Chengdu and Porto.

As shown in Table \ref{tab2}, the results demonstrate that our method achieves superior transferability across urban contexts. Possible reasons include:
1) Although AgentMove is training-free, allowing it to generalize to some extent, its reliance on semantic patterns makes it less effective in GPS reasoning tasks where explicit spatial cues are sparse.
2) Our method, on the other hand, reasons over road networks using visual maps, which more closely resembles how humans reason and generalize across unfamiliar environments. This road-aware reasoning likely underpins its consistently better generalization performance.
3) Although NextLocLLM demonstrates a certain level of cross-city transferability, it lacks knowledge of road network structures, resulting in a significant performance gap compared to the method proposed in this paper.

\subsection{Ablation Study}

\subsubsection{Ablations on Main Modules}

\begin{figure}[ht]
\begin{center}
\centerline{\includegraphics[width=0.9\columnwidth]{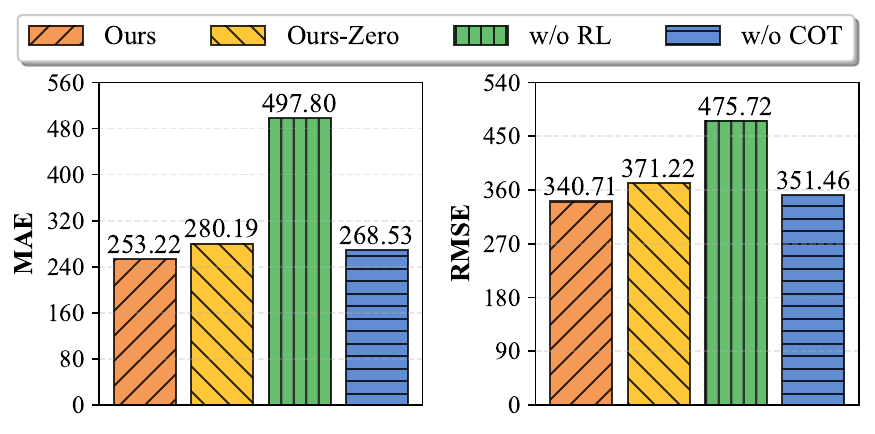}}
\caption{\textbf{Ablations on Main Modules.} The left and right figure illustrates the MAE and RMSE of each method.}
\label{figure4}
\end{center}
\end{figure}

We first conduct ablation experiments on the Chengdu dataset to assess the importance of each module. The experimental settings include:
\textbf{Ours}: The full VLMLocPredictor model.
\textbf{Ours-Zero}: RL applied directly to the base model without SFT.
\textbf{w/o RL}: Only the SFT stage is used and no RL is applied.
\textbf{w/o COT}: During the SFT stage, only the first subtask (point localization is used; the second task is removed.

As shown in Figure \ref{figure4}, the results lead to the following conclusions:
1) Our full model consistently achieves the best MAE and RMSE, validating the effectiveness of each component in the framework.
2) Interestingly, Ours-Zero also produces competitive results, suggesting that RL enables the model to autonomously learn map-based trajectory reasoning, even without explicit supervision.
3) Although w/o SFT and w/o CF cause only a slight performance drop numerically, qualitative analysis shows that the model’s reasoning outputs become incoherent or nonsensical without these SFT stages. This highlights the necessity of our proposed SFT design, particularly in shaping interpretable thinking processes.
Conclusions 2 and 3 closely mirror the findings in DeepSeek-R1, where DeepSeek-R1-Zero achieves strong results in an inexplicable manner, while the addition of CoT reasoning in DeepSeek-R1 not only maintains strong performance but also improves interpretability.

\begin{table}[h]
    \centering
    \caption{\textbf{Ablations on RL design}. We compare four different settings to demonstrate the effect of our RL design.}
    \begin{tabular}{c|cc|cc}
    \toprule
        ~ & \multicolumn{2}{|c}{\textbf{Error}} & \multicolumn{2}{|c}{\textbf{Road Dis.}} \\ 
    \midrule
        ~ & $\text{MAE}$ & $\text{RMSE}$ & MAE & RMSE \\ 
    \midrule
        Ours & \textbf{253.225}  & \textbf{340.708}  & \textbf{55.638}  & \textbf{81.371}  \\ 
        w/o R.R. & 338.917  & 514.129  & 58.780  & 87.764  \\ 
        w/ e2eSFT & 430.854  & 668.148  & 61.651  & 95.131  \\ 
        w/ Hard R.R. & 313.956  & 405.066  & 58.997  & 82.238 \\ 
    \bottomrule
    \end{tabular}
    \label{tab3}
\end{table}

\subsubsection{Ablations on design of RL}

To further evaluate the effectiveness of our RL design, we conduct additional ablation studies on the Chengdu dataset with a primary focus on the Road Reward. The settings include:
\textbf{Ours}: The complete model.
\textbf{w/o R.R.}: Road Reward is removed from our RL training.
\textbf{w/ e2eSFT}: The base model is directly SFT on the next GPS location prediction without RL.
\textbf{w/ Hard R.R.}: The continuous Road Reward is replaced with a binary version: reward is 1 if the predicted point lies on the road network, and 0 otherwise.
In this part, we additionally introduce a new metric, Road Distance, to measure the distance to the closest road. Because the baselines' road distance are rather high (MAE over 80), we only report the Road Distance in this part.

As shown in Table \ref{tab3}, the Key findings include:
1) Models trained with RL outperform SFT-only baselines, suggesting that RL offers better scalability and learning potential.
2) Incorporating the Road Reward significantly improves prediction accuracy, indicating that our design successfully encourages road-conformant reasoning.
3) The use of our soft reward function, which decays smoothly with distance from the road, outperforms the hard (0 or 1) reward. This suggests that a continuous reward signal provides a more stable and informative optimization objective for the model.  

\subsection{Case Study on Predicted Points}

\begin{figure}[ht]
\begin{center}
\centerline{\includegraphics[width=0.9\columnwidth]{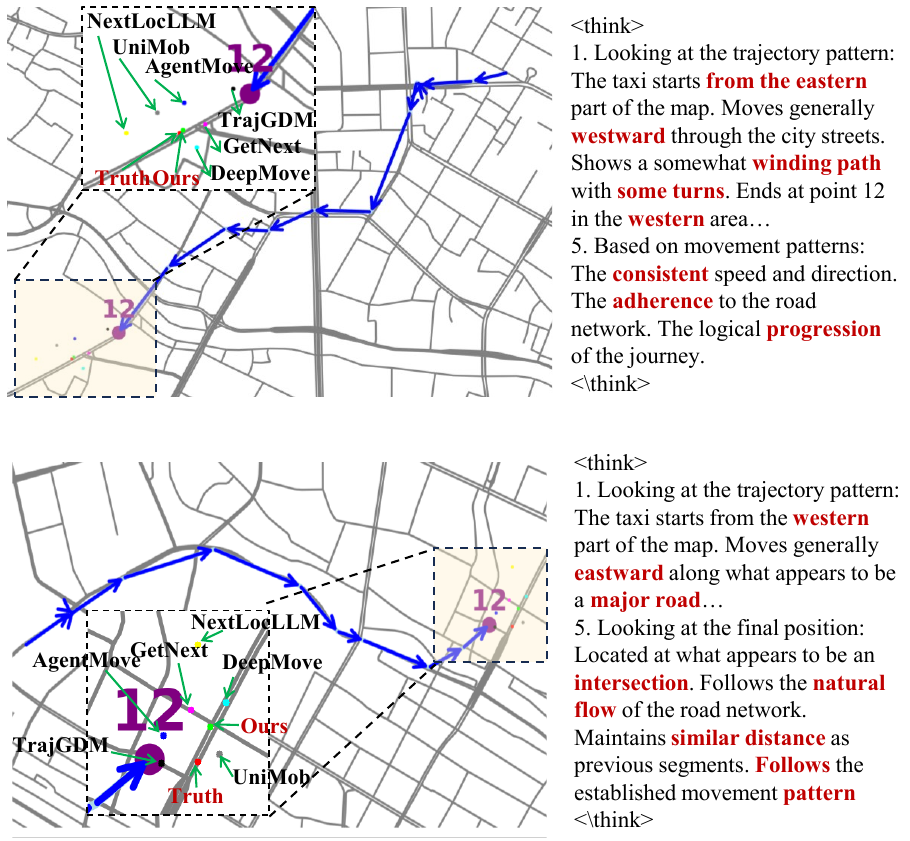}}
\caption{\textbf{Case study on predicted points.} The figure showcases two different scenarios to illustrate model behavior.}
\label{figure_case}
\end{center}
\end{figure}

To further verify the effectiveness and interpretability of our proposed method, we visualize the prediction results of various models on the Chengdu dataset. As shown in Figure~\ref{figure_case}, we present two representative scenarios.

In the upper part of the figure, we illustrate a trajectory that clearly follows a main road. And the findings are:
1) Our method accurately predicts the next GPS location, and the prediction lies naturally along the road network, forming a smooth extension of the existing trajectory. This demonstrates the spatial plausibility and effectiveness of our approach.
2) In contrast, nearly all baseline methods fail to produce predictions that align with the road network. This is likely because they rely solely on raw GPS coordinates and lack access to explicit visual or topological map information.
3) Furthermore, our model's generated reasoning text includes accurate directional cues (e.g., \textit{from the eastern}, \textit{generally westward}) and spatial descriptions (e.g., \textit{winding path}), showcasing the interpretability of our model’s predictions.

In the lower part of the figure, we present a more challenging case where the next movement requires a turn, where proceeding forward is no longer possible, and one or two turns are required, depending on the intended path.
The findings on this case are:
1) Most baseline models fail to predict this two-turn maneuver, likely due to the lack of explicit map understanding.
2) Our method successfully predicts two sequential turns and outputs a plausible next location, reflecting the precision of our spatial reasoning capability.
3) The reasoning text generated by our model also correctly identifies that the current point is at an \textit{intersection}, and recommends to \textit{follow the natural flow}, again supporting the explainability of our approach.

\subsection{Case Study on Attention Weight}

\begin{figure}[h]
\begin{center}
\centerline{\includegraphics[width=0.9\columnwidth]{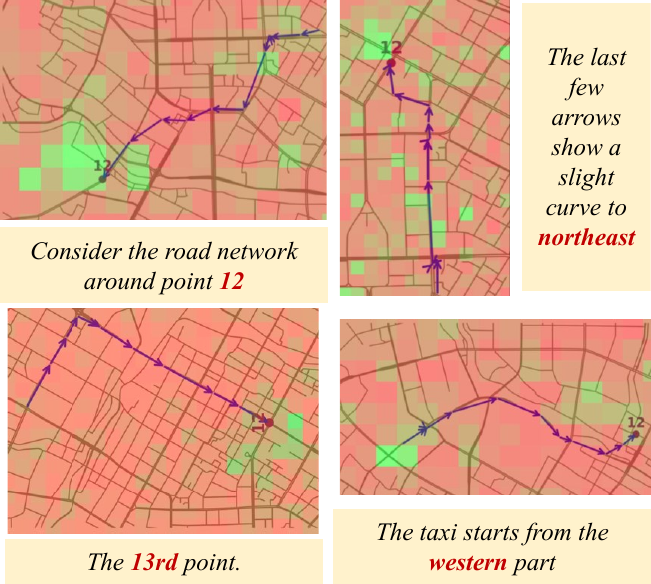}}
\caption{\textbf{Case study on Attention Weight.} The cases show our model is indeed capable of reasoning over visual maps.}
\label{figure_case2}
\end{center}
\end{figure}

To further demonstrate that our model is truly capable of reasoning over visual maps, we visualize the attention weights corresponding to the 20th layer of the VLM alongside the generated text, as shown in Figure~\ref{figure_case2}. In the visualizations, green regions indicate high attention weights, while red regions indicate lower attention. The accompanying text represents the model's generated output, and each attention heatmap corresponds to the heatmap when generating the word in red and bold.

Our observations are as follows:
1) When the model states "\textit{Consider the road network around point 12}", the attention map shows a clear focus around point 12, indicating that the model is referencing the correct spatial region.
2) When saying "\textit{The last few arrows show a slight curve to northeast}", the attention shifts to the upper-left area, which aligns with the trajectory's visual curvature.
3) While predicting point 13, the model places higher attention along the trajectory path, whereas irrelevant regions receive minimal attention.
4) When mentioning "\textit{start from the western}", the model’s attention is concentrated on the left side of the image, corresponding to the starting point of the trajectory.

The attention distributions confirm that the VLM is not generating text arbitrarily, but rather grounding its reasoning process in the visual structure of the map, ultimately leading to more accurate predictions. These examples demonstrate that, under our proposed framework, the model learns to align language with visual context.

\section{Related Work}

\subsection{Next Location Prediction}

Next Location Prediction plays a vital role in spatio-temporal modeling, with broad applications in traffic resource optimization, epidemic forecasting, and urban mobility management. This problem can be broadly categorized into two sub-tasks:
Next GPS location prediction, where the model aims to predict the exact coordinates of the next point along a trajectory.
Next POI location prediction, where the goal is to predict the next point-of-interest (POI) the user will visit.
For a long time, both tasks shared similar approaches: 

Early approaches modeled human mobility using Markov chains\cite{MarkovChains}, which capture lower-order dependencies but fail to account for longer-term sequential patterns\cite{markov1, markov2}. With the rise of deep learning, Recurrent Neural Networks(RNNs)~\cite{RNN} and Transformers~\cite{Transformer} were widely adopted for trajectory prediction tasks\cite{rnn1, rnn2, deepmove}. Later, graph-based methods and geographic feature embeddings were introduced to incorporate road network structures and spatial relationships\cite{unimob, getnext}.
Recently, LLM-based approaches such as LLM-Mob~\cite{jiang}, Agent-Move~\cite{agentmove} and NextLocLLM~\cite{nextlocllm} have been proposed to leverage textual and semantic knowledge for mobility prediction. However, these works primarily focus on next-POI-location prediction, where semantic context plays a central role. In contrast, for next GPS location prediction, most of these methods depart from how humans reason about movement that lays trajectories over a map and reason jointly.

\subsection{Vision-Language Model}

Vision-Language Models (VLMs) are designed to jointly process and reason over textual and visual inputs. The recent advances of VLMs have demonstrated strong capabilities in visual understanding and reasoning.
VLMs can be broadly categorized into two paradigms based on their learning objectives: \textbf{Contrastive Models} that do not support text generation conditioned on visual inputs such as CLIP \cite{CLIP} and GLIP \cite{GLIP}.
\textbf{Generative Models} that incorporate generative architectures, enabling generation across images and text. Examples include the GPT-4 series \cite{4o, 4o-mini}, Gemini series \cite{gemini2, geminiPro}, Pixtral \cite{pixtral12b, pixtralLarge}, and Claude series \cite{claudeHaiku, claudeSonet, Claude}. These models are trained to produce coherent, context-aware outputs conditioned on both images and text, and they have demonstrated impressive performance in tasks requiring complex reasoning. In this work, we focus specifically on generative VLMs, leveraging their ability to model intricate spatial trajectories over visual maps.

\subsection{Reinforcement Learning from Verifiable Reward}

Recently, Reinforcement Learning from Verifiable Reward (RLVR) has emerged as a key technique for enhancing large language models (LLMs). Beginning with the success of GPT-o1\cite{o1}, researchers have shown that allowing models to self-improve through interaction and feedback from the environment such as the accuracy of math problems or the pass rate of coding problems leads to substantial performance gains. This insight gave rise to models like DeepSeek-R1\cite{deepseekr1}, Gemini-Flash-Thinking\cite{gemini2}, and Claude4\cite{claude4}, all of which employ RLVR to refine model behaviors. The core RLVR techniques typically build upon policy optimization frameworks such as PPO\cite{ppo} and DPO\cite{dpo}, while recent works like DeepSeek-R1 adopt GRPO\cite{grpo}, a lightweight alternative that optimizes relative advantages without relying on value networks.
However, only recently have researchers begun to explore RL-based post-training in the visual modality. Pioneering works such as GPT-o3\cite{o3}, Visual-RFT\cite{visualrft}, and Reason-RFT\cite{reasonrft} demonstrate promising early results.  However, how to endow the success of RLVR to the next GPS location remains an open problem.

\section{Conclusion}
In this work, we explore the use of VLMs for next GPS location prediction.
We begin by introducing the Vision-Guided Location Search (VGLS) method, which evaluates a VLM’s ability to predict the next GPS location without altering its internal parameters. Our findings reveal that while larger models do exhibit an implicit capacity for next location reasoning, smaller models largely fail to demonstrate such ability.
To explicitly equip VLMs with next GPS location prediction capabilities, we propose a two-stage training pipeline.
In the first stage, we design two SFT tasks that help the VLM develop a basic understanding of our visual map and enable it to perform initial trajectory reasoning.
In the second stage, we introduce Reinforcement Learning from Visual Map Feedback with four carefully-designed rewards, which enable the model to autonomously improve its prediction ability.
Experiments on four datasets from four cities demonstrate that our proposed method achieves SOTA performance and outperforms existing LLM-based approaches in terms of cross-city generalization. 
We further show that our approach enables VLMs to explainably reason over road networks. 
In the future, our method could be extended to larger and more diverse datasets to train a general model that consistently surpasses existing approaches across different cities. 

\newpage

\newpage

\bibliography{aaai25}

\begin{thebibliography}{52}
\providecommand{\natexlab}[1]{#1}

\bibitem[{Abacha et~al.(2025)Abacha, wai Yim, Fu, Sun, Yetisgen, Xia, and Lin}]{medec}
Abacha, A.~B.; wai Yim, W.; Fu, Y.; Sun, Z.; Yetisgen, M.; Xia, F.; and Lin, T. 2025.
\newblock MEDEC: A Benchmark for Medical Error Detection and Correction in Clinical Notes.
\newblock arXiv:2412.19260.

\bibitem[{Agrawal et~al.(2024)Agrawal, Antoniak, Hanna, Bout, Chaplot, Chudnovsky, Costa, Monicault, Garg, Gervet, Ghosh, Héliou, Jacob, Jiang, Khandelwal, Lacroix, Lample, Casas, Lavril, Scao, Lo, Marshall, Martin, Mensch, Muddireddy, Nemychnikova, Pellat, Platen, Raghuraman, Rozière, Sablayrolles, Saulnier, Sauvestre, Shang, Soletskyi, Stewart, Stock, Studnia, Subramanian, Vaze, Wang, and Yang}]{pixtral12b}
Agrawal, P.; Antoniak, S.; Hanna, E.~B.; Bout, B.; Chaplot, D.; Chudnovsky, J.; Costa, D.; Monicault, B.~D.; Garg, S.; Gervet, T.; Ghosh, S.; Héliou, A.; Jacob, P.; Jiang, A.~Q.; Khandelwal, K.; Lacroix, T.; Lample, G.; Casas, D.~L.; Lavril, T.; Scao, T.~L.; Lo, A.; Marshall, W.; Martin, L.; Mensch, A.; Muddireddy, P.; Nemychnikova, V.; Pellat, M.; Platen, P.~V.; Raghuraman, N.; Rozière, B.; Sablayrolles, A.; Saulnier, L.; Sauvestre, R.; Shang, W.; Soletskyi, R.; Stewart, L.; Stock, P.; Studnia, J.; Subramanian, S.; Vaze, S.; Wang, T.; and Yang, S. 2024.
\newblock Pixtral 12B.
\newblock arXiv:2410.07073.

\bibitem[{Anthropic(2023)}]{Claude}
Anthropic. 2023.
\newblock Claude-2.1.
\newblock Accessed: 2025-01-20.

\bibitem[{Cabanas{-}Tirapu et~al.(2023)Cabanas{-}Tirapu, Dan{\'{u}}s, Moro, Sales{-}Pardo, and Guimer{\`{a}}}]{urban1}
Cabanas{-}Tirapu, O.; Dan{\'{u}}s, L.; Moro, E.; Sales{-}Pardo, M.; and Guimer{\`{a}}, R. 2023.
\newblock Human mobility is well described by closed-form gravity-like models learned automatically from data.
\newblock \emph{CoRR}, abs/2312.11281.

\bibitem[{Chen et~al.(2023)Chen, Xu, Chen, and Gao}]{rnn1}
Chen, Y.; Xu, H.; Chen, X.~M.; and Gao, Z. 2023.
\newblock A multi-scale unified model of human mobility in urban agglomerations.
\newblock \emph{Patterns}, 4(11): 100862.

\bibitem[{Chu, Zhang, and Lu(2023)}]{trajGDM}
Chu, C.; Zhang, H.; and Lu, F. 2023.
\newblock TrajGDM: {A} New Trajectory Foundation Model for Simulating Human Mobility.
\newblock In Renz, M.; and Nascimento, M.~A., eds., \emph{Proceedings of the 31st {ACM} International Conference on Advances in Geographic Information Systems, {SIGSPATIAL} 2023, Hamburg, Germany, November 13-16, 2023}, 1:1--1:2. {ACM}.

\bibitem[{Claude(2024{\natexlab{a}})}]{claudeHaiku}
Claude. 2024{\natexlab{a}}.
\newblock Claude 3 Haiku: our fastest model yet.

\bibitem[{Claude(2024{\natexlab{b}})}]{claudeSonet}
Claude. 2024{\natexlab{b}}.
\newblock Claude 3.5 Sonnet.

\bibitem[{Claude(2025)}]{claude4}
Claude. 2025.
\newblock Introducing Claude 4.

\bibitem[{Dosovitskiy et~al.(2021)Dosovitskiy, Beyer, Kolesnikov, Weissenborn, Zhai, Unterthiner, Dehghani, Minderer, Heigold, Gelly, Uszkoreit, and Houlsby}]{vit}
Dosovitskiy, A.; Beyer, L.; Kolesnikov, A.; Weissenborn, D.; Zhai, X.; Unterthiner, T.; Dehghani, M.; Minderer, M.; Heigold, G.; Gelly, S.; Uszkoreit, J.; and Houlsby, N. 2021.
\newblock An Image is Worth 16x16 Words: Transformers for Image Recognition at Scale.
\newblock In \emph{9th International Conference on Learning Representations, {ICLR} 2021, Virtual Event, Austria, May 3-7, 2021}. OpenReview.net.

\bibitem[{Elman(1990)}]{RNN}
Elman, J.~L. 1990.
\newblock Finding Structure in Time.
\newblock \emph{Cogn. Sci.}, 14(2): 179--211.

\bibitem[{Fang et~al.(2019)Fang, Zhang, Meng, Xiang, and Pan}]{res2}
Fang, S.; Zhang, Q.; Meng, G.; Xiang, S.; and Pan, C. 2019.
\newblock GSTNet: Global Spatial-Temporal Network for Traffic Flow Prediction.
\newblock In Kraus, S., ed., \emph{Proceedings of the Twenty-Eighth International Joint Conference on Artificial Intelligence, {IJCAI} 2019, Macao, China, August 10-16, 2019}, 2286--2293. ijcai.org.

\bibitem[{Feng et~al.(2025)Feng, Du, Zhao, and Li}]{agentmove}
Feng, J.; Du, Y.; Zhao, J.; and Li, Y. 2025.
\newblock AgentMove: {A} Large Language Model based Agentic Framework for Zero-shot Next Location Prediction.
\newblock In Chiruzzo, L.; Ritter, A.; and Wang, L., eds., \emph{Proceedings of the 2025 Conference of the Nations of the Americas Chapter of the Association for Computational Linguistics: Human Language Technologies, {NAACL} 2025 - Volume 1: Long Papers, Albuquerque, New Mexico, USA, April 29 - May 4, 2025}, 1322--1338. Association for Computational Linguistics.

\bibitem[{Feng et~al.(2022)Feng, Li, Lin, Rong, Sun, Guo, and Jin}]{rnn2}
Feng, J.; Li, Y.; Lin, Z.; Rong, C.; Sun, F.; Guo, D.; and Jin, D. 2022.
\newblock Context-aware Spatial-Temporal Neural Network for Citywide Crowd Flow Prediction via Modeling Long-range Spatial Dependency.
\newblock \emph{{ACM} Trans. Knowl. Discov. Data}, 16(3): 49:1--49:21.

\bibitem[{Feng et~al.(2018)Feng, Li, Zhang, Sun, Meng, Guo, and Jin}]{deepmove}
Feng, J.; Li, Y.; Zhang, C.; Sun, F.; Meng, F.; Guo, A.; and Jin, D. 2018.
\newblock DeepMove: Predicting Human Mobility with Attentional Recurrent Networks.
\newblock In Champin, P.; Gandon, F.; Lalmas, M.; and Ipeirotis, P.~G., eds., \emph{Proceedings of the 2018 World Wide Web Conference on World Wide Web, {WWW} 2018, Lyon, France, April 23-27, 2018}, 1459--1468. {ACM}.

\bibitem[{Gao et~al.(2019)Gao, Zhou, Trajcevski, Zhang, Zhong, and Zhang}]{markov1}
Gao, Q.; Zhou, F.; Trajcevski, G.; Zhang, K.; Zhong, T.; and Zhang, F. 2019.
\newblock Predicting Human Mobility via Variational Attention.
\newblock In Liu, L.; White, R.~W.; Mantrach, A.; Silvestri, F.; McAuley, J.~J.; Baeza{-}Yates, R.; and Zia, L., eds., \emph{The World Wide Web Conference, {WWW} 2019, San Francisco, CA, USA, May 13-17, 2019}, 2750--2756. {ACM}.

\bibitem[{Gemini~Team(2024{\natexlab{a}})}]{geminiPro}
Gemini~Team, G. 2024{\natexlab{a}}.
\newblock Gemini 1.5: Unlocking multimodal understanding across millions of tokens of context.

\bibitem[{Gemini~Team(2024{\natexlab{b}})}]{gemini2}
Gemini~Team, G. 2024{\natexlab{b}}.
\newblock Gemini 2.0 Flash Thinking Experimental.

\bibitem[{Guo et~al.(2025)Guo, Yang, Zhang, Song, Zhang, Xu, Zhu, Ma, Wang, Bi et~al.}]{deepseekr1}
Guo, D.; Yang, D.; Zhang, H.; Song, J.; Zhang, R.; Xu, R.; Zhu, Q.; Ma, S.; Wang, P.; Bi, X.; et~al. 2025.
\newblock Deepseek-r1: Incentivizing reasoning capability in llms via reinforcement learning.
\newblock \emph{arXiv preprint arXiv:2501.12948}.

\bibitem[{He et~al.(2016)He, Zhang, Ren, and Sun}]{resnet}
He, K.; Zhang, X.; Ren, S.; and Sun, J. 2016.
\newblock Deep Residual Learning for Image Recognition.
\newblock In \emph{2016 {IEEE} Conference on Computer Vision and Pattern Recognition, {CVPR} 2016, Las Vegas, NV, USA, June 27-30, 2016}, 770--778. {IEEE} Computer Society.

\bibitem[{Li et~al.(2022)Li, Zhang, Zhang, Yang, Li, Zhong, Wang, Yuan, Zhang, Hwang, Chang, and Gao}]{GLIP}
Li, L.~H.; Zhang, P.; Zhang, H.; Yang, J.; Li, C.; Zhong, Y.; Wang, L.; Yuan, L.; Zhang, L.; Hwang, J.; Chang, K.; and Gao, J. 2022.
\newblock Grounded Language-Image Pre-training.
\newblock In \emph{{IEEE/CVF} Conference on Computer Vision and Pattern Recognition, {CVPR} 2022, New Orleans, LA, USA, June 18-24, 2022}, 10955--10965. {IEEE}.

\bibitem[{Liu et~al.(2024)Liu, Cao, Chen, Jiang, and Cong}]{nextlocllm}
Liu, S.; Cao, N.; Chen, Y.; Jiang, Y.; and Cong, G. 2024.
\newblock nextlocllm: next location prediction using LLMs.
\newblock \emph{CoRR}, abs/2410.09129.

\bibitem[{Liu et~al.(2025)Liu, Sun, Zang, Dong, Cao, Duan, Lin, and Wang}]{visualrft}
Liu, Z.; Sun, Z.; Zang, Y.; Dong, X.; Cao, Y.; Duan, H.; Lin, D.; and Wang, J. 2025.
\newblock Visual-rft: Visual reinforcement fine-tuning.
\newblock \emph{arXiv preprint arXiv:2503.01785}.

\bibitem[{Llama(2024)}]{llama32}
Llama, M. 2024.
\newblock Llama 3.2: Revolutionizing edge AI and vision with open, customizable models.

\bibitem[{Long, Yuan, and Li(2025)}]{unimob}
Long, Q.; Yuan, Y.; and Li, Y. 2025.
\newblock A Universal Model for Human Mobility Prediction.
\newblock In Sun, Y.; Chierichetti, F.; Lauw, H.~W.; Perlich, C.; Tok, W.~H.; and Tomkins, A., eds., \emph{Proceedings of the 31st {ACM} {SIGKDD} Conference on Knowledge Discovery and Data Mining, V.1, {KDD} 2025, Toronto, ON, Canada, August 3-7, 2025}, 894--905. {ACM}.

\bibitem[{Mayemba et~al.(2024)Mayemba, Nkashama, Tshimula, Dialufuma, Muabila, Didier, Kanda, Galekwa, Fita, Mundele et~al.}]{epic3}
Mayemba, C.~N.; Nkashama, D.~K.; Tshimula, J.~M.; Dialufuma, M.~V.; Muabila, J.~T.; Didier, M.~M.; Kanda, H.; Galekwa, R.~M.; Fita, H.~D.; Mundele, S.; et~al. 2024.
\newblock A short survey of human mobility prediction in epidemic modeling from transformers to llms.
\newblock \emph{arXiv preprint arXiv:2404.16921}.

\bibitem[{MistralAI(2024)}]{pixtralLarge}
MistralAI. 2024.
\newblock Pixtral Large.

\bibitem[{Norris(1998)}]{MarkovChains}
Norris, J.~R. 1998.
\newblock \emph{Markov chains}.
\newblock 2. Cambridge university press.

\bibitem[{OpenAI(2024{\natexlab{a}})}]{4o-mini}
OpenAI. 2024{\natexlab{a}}.
\newblock GPT-4o mini: advancing cost-efficient intelligence.

\bibitem[{OpenAI(2024{\natexlab{b}})}]{4o}
OpenAI. 2024{\natexlab{b}}.
\newblock Hello GPT-4o.

\bibitem[{OpenAI(2024{\natexlab{c}})}]{o1}
OpenAI. 2024{\natexlab{c}}.
\newblock Learning to reason with LLMs.

\bibitem[{OpenAI(2025)}]{o3}
OpenAI. 2025.
\newblock Introducing OpenAI o3 and o4-mini.

\bibitem[{Pappalardo et~al.(2019)Pappalardo, Barlacchi, Pellungrini, and Simini}]{urban3}
Pappalardo, L.; Barlacchi, G.; Pellungrini, R.; and Simini, F. 2019.
\newblock Human Mobility from theory to practice: Data, Models and Applications.
\newblock In Amer{-}Yahia, S.; Mahdian, M.; Goel, A.; Houben, G.; Lerman, K.; McAuley, J.~J.; Baeza{-}Yates, R.; and Zia, L., eds., \emph{Companion of The 2019 World Wide Web Conference, {WWW} 2019, San Francisco, CA, USA, May 13-17, 2019}, 1311--1312. {ACM}.

\bibitem[{Radford et~al.(2021)Radford, Kim, Hallacy, Ramesh, Goh, Agarwal, Sastry, Askell, Mishkin, Clark, Krueger, and Sutskever}]{CLIP}
Radford, A.; Kim, J.~W.; Hallacy, C.; Ramesh, A.; Goh, G.; Agarwal, S.; Sastry, G.; Askell, A.; Mishkin, P.; Clark, J.; Krueger, G.; and Sutskever, I. 2021.
\newblock Learning Transferable Visual Models From Natural Language Supervision.
\newblock In Meila, M.; and Zhang, T., eds., \emph{Proceedings of the 38th International Conference on Machine Learning, {ICML} 2021, 18-24 July 2021, Virtual Event}, volume 139 of \emph{Proceedings of Machine Learning Research}, 8748--8763. {PMLR}.

\bibitem[{Rafailov et~al.(2023)Rafailov, Sharma, Mitchell, Manning, Ermon, and Finn}]{dpo}
Rafailov, R.; Sharma, A.; Mitchell, E.; Manning, C.~D.; Ermon, S.; and Finn, C. 2023.
\newblock Direct Preference Optimization: Your Language Model is Secretly a Reward Model.
\newblock In Oh, A.; Naumann, T.; Globerson, A.; Saenko, K.; Hardt, M.; and Levine, S., eds., \emph{Advances in Neural Information Processing Systems 36: Annual Conference on Neural Information Processing Systems 2023, NeurIPS 2023, New Orleans, LA, USA, December 10 - 16, 2023}.

\bibitem[{Schulman et~al.(2017)Schulman, Wolski, Dhariwal, Radford, and Klimov}]{ppo}
Schulman, J.; Wolski, F.; Dhariwal, P.; Radford, A.; and Klimov, O. 2017.
\newblock Proximal policy optimization algorithms.
\newblock \emph{arXiv preprint arXiv:1707.06347}.

\bibitem[{Shao et~al.(2024)Shao, Wang, Zhu, Xu, Song, Bi, Zhang, Zhang, Li, Wu et~al.}]{grpo}
Shao, Z.; Wang, P.; Zhu, Q.; Xu, R.; Song, J.; Bi, X.; Zhang, H.; Zhang, M.; Li, Y.; Wu, Y.; et~al. 2024.
\newblock Deepseekmath: Pushing the limits of mathematical reasoning in open language models.
\newblock \emph{arXiv preprint arXiv:2402.03300}.

\bibitem[{Shtedritski, Rupprecht, and Vedaldi(2023)}]{circle}
Shtedritski, A.; Rupprecht, C.; and Vedaldi, A. 2023.
\newblock What does {CLIP} know about a red circle? Visual prompt engineering for VLMs.
\newblock In \emph{{IEEE/CVF} International Conference on Computer Vision, {ICCV} 2023, Paris, France, October 1-6, 2023}, 11953--11963. {IEEE}.

\bibitem[{Song et~al.(2010)Song, Qu, Blumm, and Barab{\'a}si}]{science}
Song, C.; Qu, Z.; Blumm, N.; and Barab{\'a}si, A.-L. 2010.
\newblock Limits of predictability in human mobility.
\newblock \emph{Science}, 327(5968): 1018--1021.

\bibitem[{Tan et~al.(2025)Tan, Ji, Hao, Lin, Wang, Wang, and Zhang}]{reasonrft}
Tan, H.; Ji, Y.; Hao, X.; Lin, M.; Wang, P.; Wang, Z.; and Zhang, S. 2025.
\newblock Reason-rft: Reinforcement fine-tuning for visual reasoning.
\newblock \emph{arXiv preprint arXiv:2503.20752}.

\bibitem[{Team(2024)}]{qvq}
Team, Q. 2024.
\newblock QVQ: To See the World with Wisdom.

\bibitem[{Vaswani et~al.(2017)Vaswani, Shazeer, Parmar, Uszkoreit, Jones, Gomez, Kaiser, and Polosukhin}]{Transformer}
Vaswani, A.; Shazeer, N.; Parmar, N.; Uszkoreit, J.; Jones, L.; Gomez, A.~N.; Kaiser, L.; and Polosukhin, I. 2017.
\newblock Attention is All you Need.
\newblock In Guyon, I.; von Luxburg, U.; Bengio, S.; Wallach, H.~M.; Fergus, R.; Vishwanathan, S. V.~N.; and Garnett, R., eds., \emph{Advances in Neural Information Processing Systems 30: Annual Conference on Neural Information Processing Systems 2017, December 4-9, 2017, Long Beach, CA, {USA}}, 5998--6008.

\bibitem[{Wang et~al.(2021)Wang, Li, Jin, and Han}]{markov2}
Wang, H.; Li, Y.; Jin, D.; and Han, Z. 2021.
\newblock Attentional Markov Model for Human Mobility Prediction.
\newblock \emph{{IEEE} J. Sel. Areas Commun.}, 39(7): 2213--2225.

\bibitem[{Wang et~al.(2024{\natexlab{a}})Wang, Jiang, Yang, Wu, Onizuka, Shibasaki, Koshizuka, and Xiao}]{jiang}
Wang, J.; Jiang, R.; Yang, C.; Wu, Z.; Onizuka, M.; Shibasaki, R.; Koshizuka, N.; and Xiao, C. 2024{\natexlab{a}}.
\newblock Large Language Models as Urban Residents: An {LLM} Agent Framework for Personal Mobility Generation.
\newblock In Globersons, A.; Mackey, L.; Belgrave, D.; Fan, A.; Paquet, U.; Tomczak, J.~M.; and Zhang, C., eds., \emph{Advances in Neural Information Processing Systems 38: Annual Conference on Neural Information Processing Systems 2024, NeurIPS 2024, Vancouver, BC, Canada, December 10 - 15, 2024}.

\bibitem[{Wang et~al.(2019)Wang, Kong, Xia, and Sun}]{urban2}
Wang, J.; Kong, X.; Xia, F.; and Sun, L. 2019.
\newblock Urban Human Mobility: Data-Driven Modeling and Prediction.
\newblock \emph{{SIGKDD} Explor.}, 21(1): 1--19.

\bibitem[{Wang et~al.(2024{\natexlab{b}})Wang, Bai, Tan, Wang, Fan, Bai, Chen, Liu, Wang, Ge, Fan, Dang, Du, Ren, Men, Liu, Zhou, Zhou, and Lin}]{qwen2vl}
Wang, P.; Bai, S.; Tan, S.; Wang, S.; Fan, Z.; Bai, J.; Chen, K.; Liu, X.; Wang, J.; Ge, W.; Fan, Y.; Dang, K.; Du, M.; Ren, X.; Men, R.; Liu, D.; Zhou, C.; Zhou, J.; and Lin, J. 2024{\natexlab{b}}.
\newblock Qwen2-VL: Enhancing Vision-Language Model's Perception of the World at Any Resolution.
\newblock \emph{arXiv preprint arXiv:2409.12191}.

\bibitem[{Wu and Xie(2024)}]{saining}
Wu, P.; and Xie, S. 2024.
\newblock V*: Guided Visual Search as a Core Mechanism in Multimodal LLMs.
\newblock In \emph{{IEEE/CVF} Conference on Computer Vision and Pattern Recognition, {CVPR} 2024, Seattle, WA, USA, June 16-22, 2024}, 13084--13094. {IEEE}.

\bibitem[{Yang, Liu, and Zhao(2022)}]{getnext}
Yang, S.; Liu, J.; and Zhao, K. 2022.
\newblock GETNext: Trajectory Flow Map Enhanced Transformer for Next {POI} Recommendation.
\newblock In Amig{\'{o}}, E.; Castells, P.; Gonzalo, J.; Carterette, B.; Culpepper, J.~S.; and Kazai, G., eds., \emph{{SIGIR} '22: The 45th International {ACM} {SIGIR} Conference on Research and Development in Information Retrieval, Madrid, Spain, July 11 - 15, 2022}, 1144--1153. {ACM}.

\bibitem[{Zhang and Patras(2018)}]{res3}
Zhang, C.; and Patras, P. 2018.
\newblock Long-Term Mobile Traffic Forecasting Using Deep Spatio-Temporal Neural Networks.
\newblock In \emph{Proceedings of the Nineteenth {ACM} International Symposium on Mobile Ad Hoc Networking and Computing, MobiHoc 2018, Los Angeles, CA, USA, June 26-29, 2018}, 231--240. {ACM}.

\bibitem[{Zhang et~al.(2025{\natexlab{a}})Zhang, Yang, Wu, and Li}]{res1}
Zhang, J.; Yang, Y.; Wu, X.; and Li, S. 2025{\natexlab{a}}.
\newblock Spatio-temporal transformer and graph convolutional networks based traffic flow prediction.
\newblock \emph{Scientific Reports}, 15(1): 24299.

\bibitem[{Zhang et~al.(2025{\natexlab{b}})Zhang, Tai, Yao, Yang, Ruggeri, Shaman, and Pei}]{epic1}
Zhang, R.; Tai, J.; Yao, Q.; Yang, W.; Ruggeri, K.; Shaman, J.; and Pei, S. 2025{\natexlab{b}}.
\newblock Behavior-driven forecasts of neighborhood-level COVID-19 spread in New York City.
\newblock \emph{PLOS Computational Biology}, 21(4): e1012979.

\bibitem[{Zhao et~al.(2024)Zhao, Wang, Zhang, Shi, Xu, Man, Chen, Li, Li, Sun et~al.}]{epic2}
Zhao, B.; Wang, X.; Zhang, T.; Shi, R.; Xu, F.; Man, F.; Chen, E.; Li, Y.; Li, Y.; Sun, T.; et~al. 2024.
\newblock Estimating and modeling spontaneous mobility changes during the COVID-19 pandemic without stay-at-home orders.
\newblock \emph{Humanities and Social Sciences Communications}, 11(1): 1--15.

\end{thebibliography}

\clearpage

\newpage

\begin{figure*}[h]
\begin{center}
\centerline{\includegraphics[width=2\columnwidth]{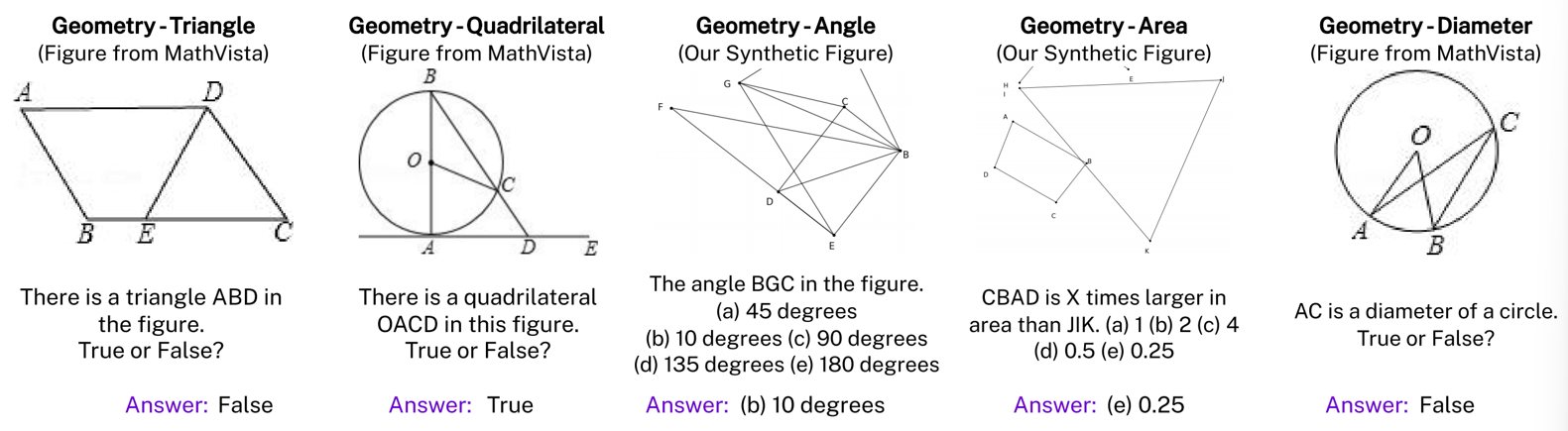}}
\caption{Illustration of the VisOnlyQA dataset.}
\label{figVis}
\end{center}
\end{figure*}

\section{Appendix}

\begin{figure*}[t]
\begin{center}
\centerline{\includegraphics[width=2.1\columnwidth]{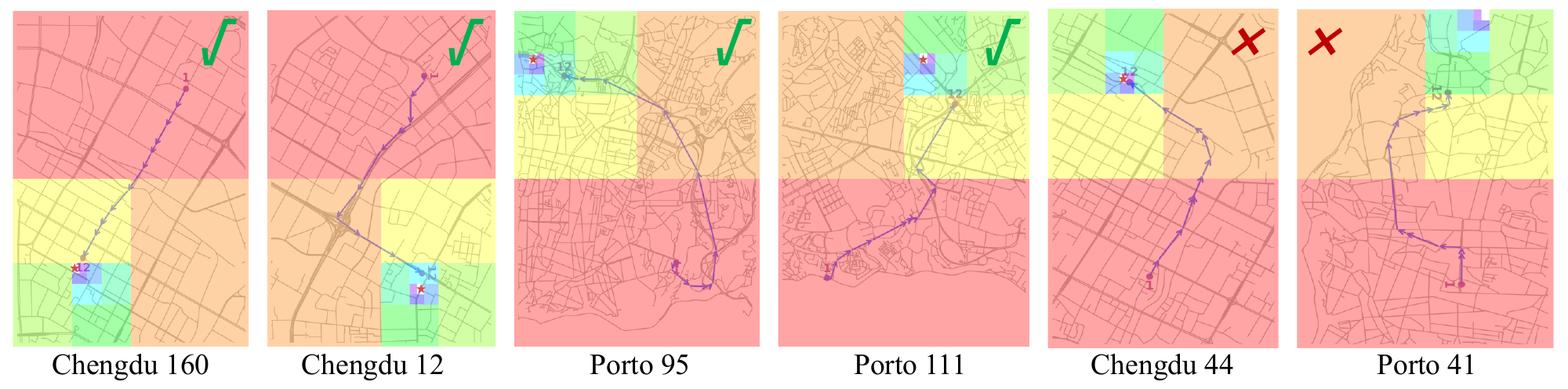}}
\caption{\textbf{Case Study of the Preliminary Experiment}. The examples are all taken from the predictions of Claude3.5 Sonet. The first four represent reasonable predictions, while the last two correspond to less reasonable predictions. The colored regions indicate areas that the model has excluded from consideration, and the red star marks the model's final predicted location.}
\label{fig_casP}
\end{center}
\end{figure*}

\subsection{Experimental Settings of the Preliminary Experiment}

In this section, we introduce additional experimental settings of the \textbf{preliminary task}. Generally, the preliminary task uses the same Chengdu dataset as the main experiment. We only report the baselines and additional experimental settings in the preliminary experiment here.

\subsubsection{Experimental Settings.} Most vision-language models were accessed via the OpenRouter API\footnote{\url{https://openrouter.ai}}. Especially, GeminiFlash2 Thinking was accessed through Google AI Studio\footnote{\url{https://aistudio.google.com/}}.

\subsubsection{Calculation of MAE and RMSE}

To evaluate the performance of our next GPS location prediction model, we use two widely adopted distance-based metrics: \textbf{Mean Absolute Error (MAE)} and \textbf{Root Mean Squared Error (RMSE)}. 

\textbf{Mean Absolute Error (MAE)}. MAE measures the average absolute Euclidean distance between predicted and ground-truth coordinates. It is defined as:

\begin{equation}
\text{MAE} = \frac{1}{N} \sum_{i=1}^{N} \left( \sqrt{ (\hat{x}_i - x_i)^2 + (\hat{y}_i - y_i)^2  }\right)
\end{equation}
\textbf{ Root Mean Squared Error (RMSE)}. RMSE emphasizes larger errors by squaring the differences before averaging. It is defined as:

\begin{equation}
\text{RMSE} = \sqrt{ \frac{1}{N} \sum_{i=1}^{N} \left\|  \sqrt{ (\hat{x}_i - x_i)^2 + (\hat{y}_i - y_i)^2  } \right\|_2^2 }
\end{equation}


\subsubsection{VLM Baselines.} In the preliminary experiment, we only compare with general VLMs: including GeminiFlash2 Thinking 1219\cite{gemini2} (Guessed to be $\sim$175B), and QVQ-72B-Preview\cite{qvq}, Claude3.5-Sonet ($\sim$175B)\cite{claudeSonet}, Claude3-Haiku (Guessed to be $\sim$8B)\cite{claudeHaiku}, Qwen2VL-72B-Instruct\cite{qwen2vl}, Qwen2VL-7B-Instruct, Qwen2VL-2B-Instruct\cite{qwen2vl}, Pixtral-Large-2411(124B)\cite{pixtralLarge}, Pixtral-12B\cite{pixtral12b}, GPT4o ($\sim$200B)\cite{4o}, GPT4o-mini ($\sim$8B)\cite{4o-mini}, Llama3.2-90B-vision-Instruct\cite{llama32}, and Llama3.2-11B-vision-Instruct\cite{llama32}. The number of parameters of some models is derived from \cite{medec}. The \textit{Instruct} in the model name may be omitted for simplicity. Because GPT-o1\cite{o1} temporarily only supports their tier-5 users, we did not compare with it.

\subsection{Further Descriptions of Each Dataset}

We utilize four publicly available datasets collected from Chengdu\footnote{\url{http://www.dcjingsai.com}}, Porto\footnote{\url{https://www.kaggle.com/crailtap/taxi-trajectory}}, San Francisco\footnote{\url{http://crawdad.org/epfl/mobility/20090224/}}, and Rome\footnote{\url{http://crawdad.org/roma/taxi/20140717/}}. Due to computational constraints, we randomly sample 1,500 trajectories from each dataset and split them into training, validation, and test sets using a 7:1:2 ratio.

In the SFT stage for point localization, we generate 12 question-answer pairs for each trajectory, corresponding to the location of each point on the map. 

In the SFT stage for Chain-of-Thought (CoT) generation, we curate 300 high-quality CoT samples per dataset. These samples are used to train the model's basic reasoning ability, and some of the examples are shown in later sections.

\subsection{The Predictability of Next Location}

While the next location inherently involves some degree of randomness, one might question whether it is truly predictable.

However, from an intuitive standpoint, the next location is far from arbitrary. In the context of urban mobility, especially in taxi trajectories, drivers often follow near-optimal routes constrained by the road network and influenced by practical factors such as traffic flow and passenger destinations. Crucially, these constraints and patterns are often reflected in the driver’s past trajectory, including route choice, directionality, and speed. As a result, the feasible space for the next point is significantly narrowed, making the next location partially deterministic and thus predictable to a meaningful extent.

This intuition is supported by empirical findings. For example, \cite{science} demonstrates—through rigorous mathematical analysis—that human mobility, as captured by signaling data, exhibits a predictability upper bound of nearly 70\%. This confirms that the movement patterns contain strong patterns.

Moreover, even if the next location cannot be predicted with perfect accuracy, the task remains valuable and well-founded. It shares similarities with next-token prediction in language modeling, where multiple plausible continuations exist, yet predictive modeling has led to breakthroughs in understanding and generation. By analogy, next location prediction can be framed as a probabilistic and context-aware spatial reasoning task, making it a meaningful and well-posed challenge.

\subsection{Out-of-Distribution Evaluation}

\begin{table}[h]
    \centering
    \caption{\textbf{Comparison with Vision Encoder Baselines}}
    \begin{tabular}{c|ccccc}
    \toprule
        ~ & Triangle & Quadrilateral & Angle & Area & Diameter \\ 
    \midrule
        Random & 50\%  & 50\%  & 20\%  & 20\% & 50\%    \\ 
        Base & 44\%  & 44\%  & 19\%  & 23\% & 45\%   \\ 
        Ours & 48\%  & 45\%  & 29\%  & 29\% & 46\%     \\ 
    \bottomrule
    \end{tabular}
    \label{tabood}
\end{table}

To assess the broader applicability of our method, we conducted experiments on the VisOnlyQA dataset. We selected five types of geometry-related questions: identifying triangles, identifying quadrilaterals, judging angles, comparing areas, and determining diameters. Example questions are shown in the Figure \ref{figVis}. We compared random prediction, Qwen2-VL, and our VLMLocPredictor trained solely on the Chengdu dataset.

As shown in the Table \ref{tabood}, the results indicate:
1. Our method consistently outperforms Qwen2-VL, demonstrating out-of-distribution generalized ability. Notably, in the Angle and Area tasks, our model shows significant gains over both the baseline and random prediction. This may be because reasoning about angles and spatial regions is a core skill for next location prediction tasks.
2. Although improvements on the other three tasks are less pronounced, our method still steers the model toward more reasonable predictions compared to random guessing.

\subsection{Comparison with Vision Encoder Baselines}

\begin{table}[h]
    \centering
    \caption{\textbf{Comparison with Vision Encoder Baselines}}
    \begin{tabular}{c|cc}
    \toprule
        ~ & $\text{MAE}$ & $\text{RMSE}$ \\ 
    \midrule
        ResNet & 894.873  & 1102.950   \\ 
        ViT & 1186.355  & 1270.835 \\ 
        Ours & 253.225  & 340.708    \\ 
    \bottomrule
    \end{tabular}
    \label{tabVEB}
\end{table}

To evaluate the difference between our method and purely vision-based approaches, we compared our model against representations obtained from ResNet\cite{resnet} and ViT\cite{vit}, followed by linear probing. The results in Table \ref{tabVEB} show that these vision encoders, likely due to their lack of exposure to road network data during pretraining, perform significantly worse than our proposed method and other baselines. Therefore, this experiment was only conducted on the Chengdu dataset and was not included in the main paper.

\subsection{Case Study of the Preliminary Experiment}

In this section, we present a visual analysis of selected prediction results from Claude 3.5 Sonet in the Figure \ref{fig_casP}. The first four examples illustrate reasonable predictions, while the last two exhibit less reasonable outcomes.

First, in the Chengdu160 trajectory, which features near-uniform motion, the model accurately predicts the next location along the uniform trajectory, demonstrating its inherent spatial reasoning capabilities. Similarly, in the Chengdu12 trajectory, the model successfully identifies a feasible turning point after a curve, suggesting that it has learned a reasonable understanding of the topological relationships within the road network.
In more complex scenarios, such as Porto95 and Porto111, the model effectively selects a viable next location based on both the trajectory and road network constraints, further supporting its capability for next location prediction. 

However, the model's predictions can sometimes be influenced by previous trajectory patterns. For instance, in Chengdu44, the movement from the fourth to the fifth point is minimal, leading the model to predict a stop at the 13th point. Conversely, in Porto41, where the displacement between the 6th and 7th points is large, the model overestimates the movement range when predicting the 13th point.
While these latter predictions may appear suboptimal, they still indicate that the model is capturing historical trajectory patterns. This suggests that its predictive capabilities can be further improved with additional refinement and training.

\subsection{Increasing the Number of Input Points}

\begin{figure}[h]

\begin{center}
\centerline{\includegraphics[width=\columnwidth]{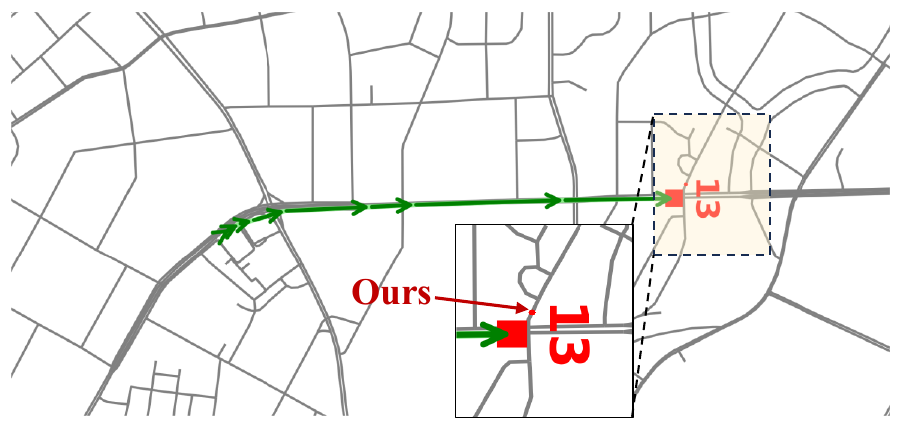}}
\caption{Case Study of Increasing the Number of Input Points.}
\label{figInc}
\end{center}

\end{figure}

To demonstrate the generalization capability of our proposed method, we examine whether the model can predict the 14th point based on the first 13 points. As the visualization results shown in Figure \ref{figInc}, the model is able to generate a reasonable prediction for the 14th location, indicating strong extensibility. Notably, while most prior deep learning models operate with a fixed context length, our model—trained only with 12-point inputs—successfully generalizes to settings with 13-point inputs. This highlights the robustness and effectiveness of our approach.

\subsection{Prompts}

As the Qwen-2-VL has basic object detection ability, and can output object detection result in the format like  $<|object\_ref\_start|>the 1th point<|object\_ref\_end|><|box\_start|>(749,500),(750,501)<|box\_end|>$, we reuse their ability in our method. 

\begin{tcolorbox}[colback=blue!3!white,
                  colframe=blue!75!black,
                  title=Prompot of VGLS,
                  fonttitle=\bfseries,
                  coltitle=white,
                  sharp corners=south,
                  breakable]

\textbf{Hello, you are an expert in next-location prediction.}

\textbf{Known Information:}

\begin{itemize} 
\item  1. This trajectory represents the movement of a taxi driver.
\item  2. Each arrow represents the distance and direction of movement over approximately 45 seconds
\item  3. The 11 arrows in the diagram show the sequential movement from the first trajectory point to the 12th trajectory point.
\item  4. The starting position is marked with a purple dot labeled as point 1 in the diagram, and the current position is marked with a purple dot labeled as point 12.
\item  5. Gray lines in the diagram represent drivable roads, while white areas indicate buildings or other obstacles.
\end{itemize}

\textbf{Question:}
\begin{itemize} 
\item  Based on the following requirements, determine whether the next position (the 13th trajectory point approximately 45 seconds later from the current purple dot position (the 12th trajectory point) is more likely to be in the blue region or the yellow region.
\end{itemize}

\textbf{Requirements:}

\begin{itemize} 
  \item Return the answer in JSON format, containing a key ANS. If the final answer is the blue region, the value should be 0; if the final answer is the yellow region, the value should be 1.2. Let's think step by step.
\end{itemize}
\end{tcolorbox}

\begin{tcolorbox}[colback=blue!3!white,
                  colframe=blue!75!black,
                  title=Prompt of Point Localization,
                  fonttitle=\bfseries,
                  coltitle=white,
                  sharp corners=south,
                  breakable]

\textbf{You are a trajectory reasoning assistant, and you need to complete the trajectory reasoning task according to the following rules.} \\

\textbf{Known Information:}

\begin{itemize} 
\item 1. This trajectory represents the movement of a taxi driver. 
\item 2. Each arrow represents the distance and direction of movement over approximately 45 seconds. 
\item 3. The 11 arrows in the diagram show the sequential movement from the first trajectory point to the 12th trajectory point. 
\item 4. The starting position is marked with a purple dot labeled as point 1 in the diagram, and the last position is marked with a purple dot labeled as point 12.
\item 5. Gray lines in the diagram represent drivable roads, while white areas indicate buildings or other obstacles. 
\end{itemize}

\textbf{Requirements:}
\begin{itemize} 
  \item 1. Let's think step by step. 
  \item 2. Output the final answer in <answer> </answer> tags.
  \item3. The final answer should be covered in a $<|box\_start|> <|box\_end|>$ tags. For example, $<|box\_start|>(749,500),(750,501)<|box\_end|>$. 
\end{itemize}

\textbf{Questions:}
\begin{itemize} 
    \item 1. Based on the following requirements, output the location of the 1th trajectory point. . In the format like $<|object\_ref\_start|>the 1th point<|object\_ref\_end|><|box\_start|>(749,500),(750,501)<|box\_end|>$
\end{itemize}

$<image>$
\end{tcolorbox}

\begin{tcolorbox}[colback=blue!3!white,
                  colframe=blue!75!black,
                  title=Prompt of Generating the COT Data,
                  fonttitle=\bfseries,
                  coltitle=white,
                  sharp corners=south,
                  breakable]

\textbf{You are a trajectory reasoning assistant, and you need to complete the trajectory reasoning task according to the following rules.} \\

\textbf{Known Information:}

\begin{itemize} 
\item 1. This trajectory represents the movement of a taxi driver. 
\item 2. Each arrow represents the distance and direction of movement over approximately 45 seconds. 
\item 3. The 12 arrows in the diagram show the sequential movement from the first trajectory point to the 12th trajectory point. 
\item 4. The starting position is marked with a purple dot labeled as point 1 in the diagram, and the last position is marked with a purple dot labeled as point 13.
\item 5. Gray lines in the diagram represent drivable roads, while white areas indicate buildings or other obstacles. 
\end{itemize}

\textbf{Requirements:}
\begin{itemize} 
  \item 1. Let's think step by step. 
  \item 2. Output a confidence score between 0 and 1 after your thinking process in json format like {"confidence": 0.9}.
\end{itemize}

\textbf{Questions:}
\begin{itemize} 
\item 1. You should generate a chain-of-thought thinking process to reason the 13th point without mentioning it. And do not conclude, just output reasoning process.
\item 2. You should last judge whether the 13th point is the current position of the taxi driver given the first 12 points, output the confidence of your judgment.

\end{itemize}

\end{tcolorbox}

\begin{tcolorbox}[colback=blue!3!white,
                  colframe=blue!75!black,
                  title=Prompt of Predicting the 13th point,
                  fonttitle=\bfseries,
                  coltitle=white,
                  sharp corners=south,
                  breakable]

\textbf{You are a trajectory reasoning assistant, and you need to complete the trajectory reasoning task according to the following rules.} \\

\textbf{Known Information:}

\begin{itemize} 
\item 1. This trajectory represents the movement of a taxi driver. 
\item 2. Each arrow represents the distance and direction of movement over approximately 45 seconds. 
\item 3. The 11 arrows in the diagram show the sequential movement from the first trajectory point to the 12th trajectory point. 
\item 4. The starting position is marked with a purple dot labeled as point 1 in the diagram, and the last position is marked with a purple dot labeled as point 12.
\item 5. Gray lines in the diagram represent drivable roads, while white areas indicate buildings or other obstacles. 
\end{itemize}

\textbf{Requirements:}
\begin{itemize} 
  \item 1. Let's think step by step. 
  \item 2. Output the final answer in <answer> </answer> tags.
  \item3. The final answer should be covered in a $<|box\_start|> <|box\_end|>$ tags. For example, $<|box\_start|>(749,500),(750,501)<|box\_end|>$. 
\end{itemize}

\textbf{Questions:}
\begin{itemize} 
    \item 1. Based on the following requirements, predict the location of the 13th trajectory point. In the format like $<|object\_ref\_start|>the 13th point<|object\_ref\_end|><|box\_start|>(749,500),(750,501)<|box\_end|>$
\end{itemize}

$<image>$
\end{tcolorbox}

\subsection{Sample Outputs of VLMLocPredictor}

We omit the $<|object\_ref\_start|>, <|object\_ref\_end|>, <|box\_start|>, <|box\_end|>$ symbols in the output. In addition, Qwen-2-VL scale the output position to 1000 x 1000, which means the output coordinate need to be rescaled to original position.

\begin{tcolorbox}[colback=blue!3!white,
                  colframe=blue!75!black,
                  title=Sample1,
                  fonttitle=\bfseries,
                  coltitle=white,
                  sharp corners=south,
                  breakable]

\begin{center}
\includegraphics[width=0.8\linewidth]{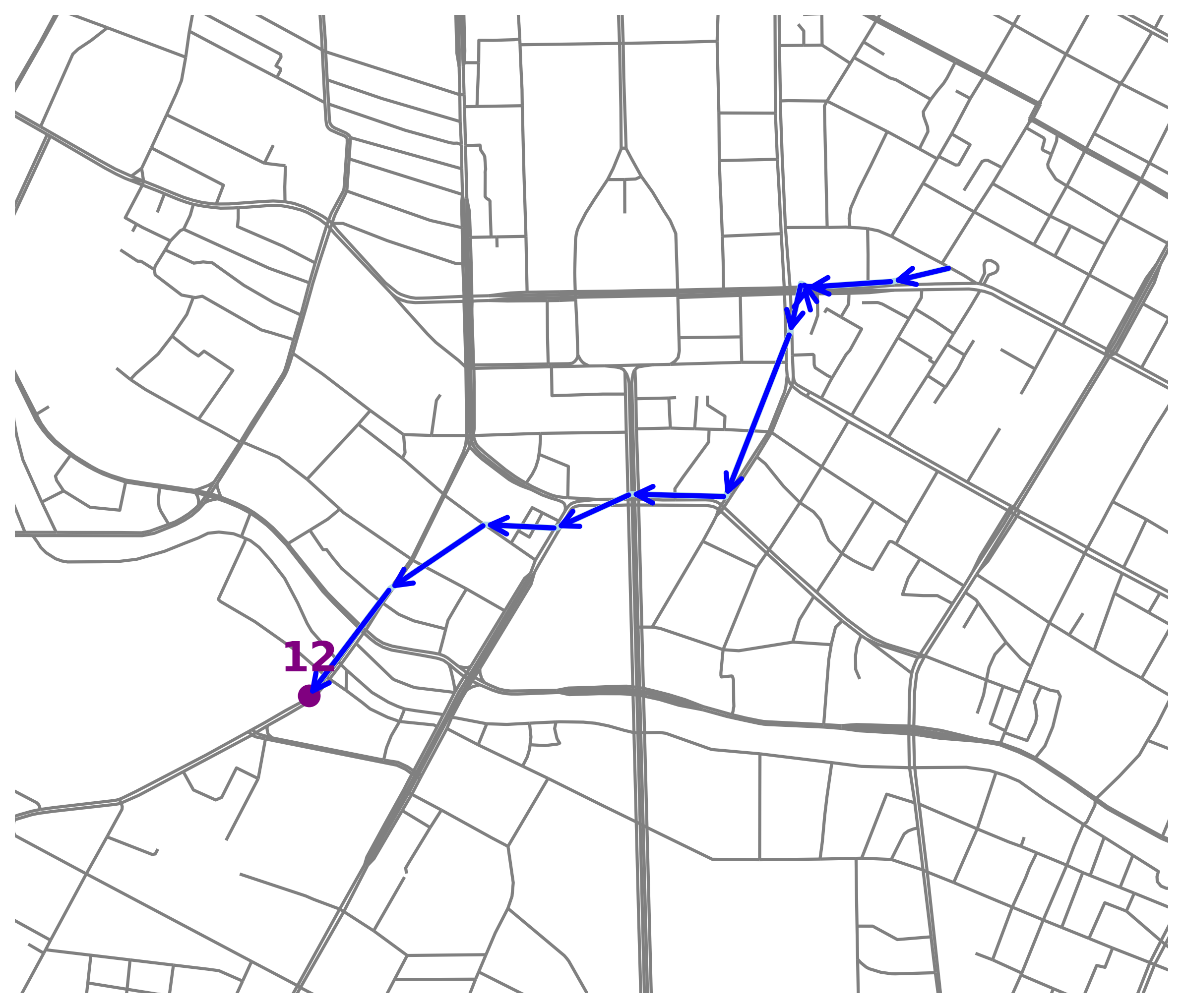} 
\end{center}

$<think>$ \\
Let me reason about this step by step:

\textbf{1. Looking at the trajectory pattern:}
\begin{itemize} 
\item - The taxi starts from the eastern part of the map
\item - Moves generally westward through the city streets
\item - Shows a somewhat winding path with some turns
\item - Ends at point 12 in the western area
\end{itemize}

\textbf{2. Analyzing the movement characteristics:}
\begin{itemize} 
\item - Each arrow represents about 45 seconds of movement
\item  - The arrows are fairly consistent in length, suggesting relatively constant speed
\item - The driver follows the available road network
\item - No illegal movements through buildings or obstacles are shown
\end{itemize}

\textbf{3. Examining the final movements:}
\begin{itemize} 
\item  - The last few arrows (9-12) show a clear westward movement
\item - The driver appears to be following a main road
\item - The final arrows lead to point 12
\item - The movement pattern suggests a typical taxi service behavior
\end{itemize}

\textbf{4. Considering the road network around point 12:}
\begin{itemize} 
\item - Point 12 is located at what appears to be an intersection
\item - There are multiple possible directions available from this point
\item - The previous movement pattern suggests a likely destination rather than a random stop
\end{itemize}

\textbf{5. Based on movement patterns:}
\begin{itemize} 
\item - The consistent speed and direction
\item - The adherence to the road network
\item - The logical progression of the journey
\item - The typical behavior of taxi services
\end{itemize}

$</think>$\\
$<answer>$ \\ 
the 13th point(199,730),(200,731) \\
$</answer>$\\ 

\end{tcolorbox}

\begin{tcolorbox}[colback=blue!3!white,
                  colframe=blue!75!black,
                  title=Sample2,
                  fonttitle=\bfseries,
                  coltitle=white,
                  sharp corners=south,
                  breakable]

\begin{center}
\includegraphics[width=0.8\linewidth]{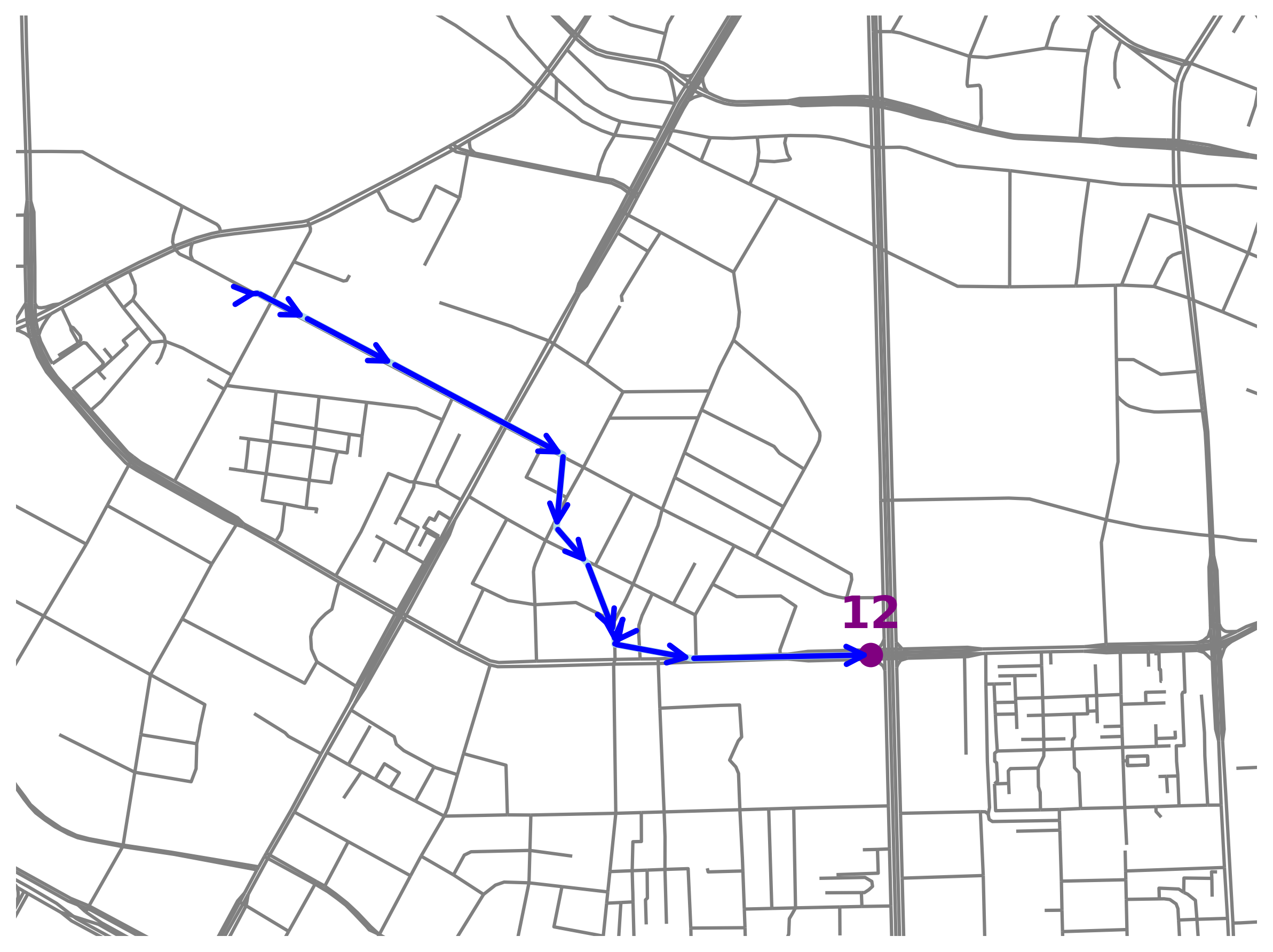} 
\end{center}

$<think>$ \\
Let me reason about this step by step:

\textbf{1. Looking at the trajectory pattern:}
\begin{itemize}
\item - The taxi starts from the western part of the map
\item - Moves generally eastward along what appears to be a major road
\item - The movement is relatively consistent in speed (arrow lengths are similar)
\item - The final arrows lead to point 12 in the eastern part of the map

\end{itemize}

\textbf{2. Analyzing the movement characteristics:}
\begin{itemize}
\item - Each arrow represents about 45 seconds of movement
\item - The driver maintains a fairly steady speed throughout
\item - No sudden turns or speed changes are observed
\item - The movement follows the available road network

\end{itemize}

\textbf{3. Examining the final segment:}
\begin{itemize}
\item - The last few arrows (9-12) show a slight curve to the southeast
\item - The speed remains consistent
\item - The driver appears to be following the main road
\item - The final arrows lead to point 12

\end{itemize}

\textbf{4. Considering the road network:}
\begin{itemize}
\item - The area around point 12 shows multiple possible road options
\item - The previous movement pattern suggests a preference for main roads
\item - The available road network constrains possible movement directions
\item - The main road continues in the same general direction

\end{itemize}

\textbf{5. Based on movement patterns:}
\begin{itemize} 
\item - The consistent speed and direction suggest a predictable trajectory
\item - No erratic movements or unexpected turns
\item - The driver appears to be following a planned route
\item - The final arrows indicate a natural continuation of the previous movement
\end{itemize}

$</think>$\\
$<answer>$ \\ 
the 13th point(790,680),(791,681) \\
$</answer>$\\ 

\end{tcolorbox}

\begin{tcolorbox}[colback=blue!3!white,
                  colframe=blue!75!black,
                  title=Sample3,
                  fonttitle=\bfseries,
                  coltitle=white,
                  sharp corners=south,
                  breakable]

\begin{center}
\includegraphics[width=0.6\linewidth]{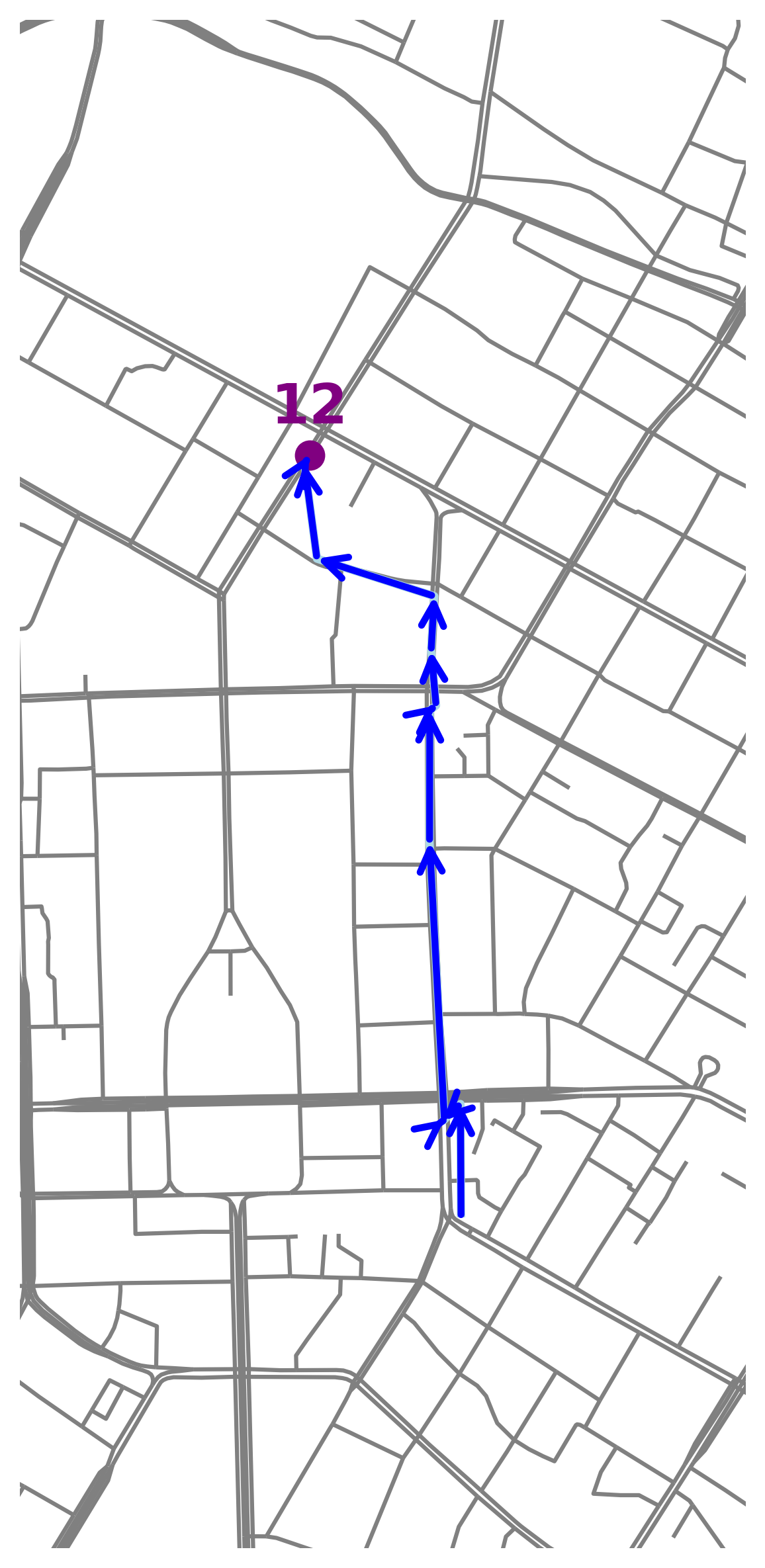} 
\end{center}

$<think>$ \\
Let me reason about this step by step:

\textbf{1. Looking at the trajectory pattern:}
\begin{itemize} 
\item - The taxi starts from the bottom of the map
\item - Moves generally northward along what appears to be a major road
\item - The arrows show consistent speed and direction for most segments

\end{itemize}

\textbf{2. Analyzing the movement characteristics:}
\begin{itemize} 
\item - Each arrow represents about 45 seconds of movement
\item - The arrows are roughly equal in length, suggesting relatively constant speed
\item - The driver follows the road network without any sudden turns

\end{itemize}

\textbf{3. Examining the final Movements:}
\begin{itemize} 
\item - The last few arrows (9-12) show a slight curve to the northeast
\item - The speed appears consistent throughout
\item - The driver follows the available road network

\end{itemize}

\textbf{4. Considering the road network around point 12:}
\begin{itemize} 
\item - There are limited possible directions for the next movement
\item \item - The previous movement pattern suggests a preference for following the main road
\item - The available road options are constrained by the road layout

\end{itemize}

\textbf{5. Based on movement patterns:}
\begin{itemize} 
\item - The consistent speed and direction suggest a predictable trajectory
\item \item - The driver appears to be following a main arterial route
\item - The final movements indicate a slight curve rather than a sharp turn

\end{itemize}

\textbf{6. Analyzing the final position (point 13):}
\begin{itemize} 
\item - Located slightly northeast of point 12
\item - Follows the available road network
\item - Maintains similar distance as previous segments
\item - Follows the natural flow of the previous movement pattern
\end{itemize}

$</think>$\\
$<answer>$ \\ 
the 13th point(380,259),(381,260)\\
$</answer>$\\ 

\end{tcolorbox}

\end{document}